\documentclass[preprint]{elsarticle}

\usepackage{hyperref}
\usepackage{times}
\usepackage{epsfig}
\usepackage{graphicx}
\usepackage{amsmath}
\usepackage{amssymb}
\usepackage{multirow}
\usepackage{caption}
\usepackage{subcaption}
\usepackage{algorithm2e}
\usepackage{algorithmic}

\journal{Published in Pattern Recognition, \href{https://doi.org/10.1016/j.patcog.2017.04.017}{https://doi.org/10.1016/j.patcog.2017.04.017}, Elsevier.}

\begin{document}

\begin{frontmatter}

\title{Estimating 3D Trajectories from 2D Projections via Disjunctive Factored Four-Way Conditional Restricted Boltzmann Machines}


\author[tue]{Decebal Constantin Mocanu \corref{mycorrespondingauthor}}
\ead{d.c.mocanu@tue.nl}
\author[aub]{Haitham Bou Ammar}
\author[uow]{Luis Puig}
\author[pen]{Eric Eaton}
\author[tue]{Antonio Liotta}

\cortext[mycorrespondingauthor]{Corresponding author}

\address[tue]{Department of Electrical Engineering, Eindhoven University of Technology, Eindhoven, the Netherlands.}
\address[aub]{Department of Computer Science, American University of Beirut, Beirut, Lebanon.}
\address[uow]{Department of Computer Science \& Engineering, University of Washington, Seattle, USA.}
\address[pen]{Department of Computer and Information Science, University of Pennsylvania, Philadelphia, USA.}


\begin{abstract}
Estimation, recognition, and near-future prediction of 3D trajectories based on their two dimensional projections available from one camera source is an exceptionally difficult problem due to uncertainty in the trajectories and environment, high dimensionality of the specific trajectory states, lack of enough labeled data and so on. In this article, we propose a solution to solve this problem based on a novel deep learning model dubbed \textit{disjunctive factored four-way conditional restricted Boltzmann machine} (DFFW-CRBM). Our method improves state-of-the-art deep learning techniques  for high dimensional time-series modeling by introducing a novel tensor factorization capable of driving forth order Boltzmann machines to considerably lower energy levels, at no computational costs. DFFW-CRBMs are capable of accurately estimating, recognizing, and performing near-future prediction of three-dimensional trajectories from their 2D projections while requiring limited amount of labeled data. We evaluate our method on both simulated and real-world data, showing its effectiveness in predicting and classifying complex ball trajectories and human activities. 
\end{abstract}


\end{frontmatter}


\section{Introduction}
Estimating and predicting trajectories in three-dimensional spaces based on two-dimensional projections available from \emph{one} camera source is an open problem with wide-ranging applicability including entertainment~\cite{3dtrajrobotics}, medicine~\cite{3dtrajmedicie}, biology ~\cite{3dtrajbiology},  physics~\cite{3dtrajdarkmatter}, etc. Unfortunately, solving this problem is exceptionally difficult due to a variety of challenges, such as the variability of states of the trajectories, partial occlusions due to self articulation and layering of objects in the scene, and the loss of 3D information resulting from observing trajectories through 2D planar image projections. A variety of techniques have considered variants of this problem by incorporating additional sensors, e.g., cameras~\cite{3dtrajwith3cameras}, radars~\cite{3dtrajwithradar}, which provide new data for geometric solvers allowing for accurate estimation and prediction. Though compelling, the success of these methods arrives at increased costs (e.g., incorporating new sensors) and computational complexities (e.g., handling more inputs geometrically). 

The problem above, however, can be framed as a time-series estimation and prediction one, for which numerous machine learning algorithms can be applied. An emerging trend in machine learning for computer vision and pattern recognition is deep learning (DL) which has been successfully applied in a variety of fields, e.g., multi-class classification~\cite{Larochelle+Bengio-2008}, collaborative filtering~\cite{Salakhutdinov07restrictedboltzmann}, image quality assessment~\cite{mocanu2014deep}, reinforcement learning~\cite{mnih15}, transfer learning~\cite{ecml2013dec}, information retrieval~\cite{Gehler06therate}, depth estimation~\cite{3ddepthnips2014}, face recognition~\cite{escalerafacerecognition}, and activity recognition~\cite{escaleraactivityrecognition}. Most related to this work are \emph{temporal-based} deep learners, e.g.,~\cite{Shotton13, temporalrbm}, which we briefly review next. Extending on standard restricted Boltzmann machines (RBMs)\cite{originalrbm}, temporal RBMs (TRBMs) consider a succession of RBMs, one for each time frame, allowing them to perform accurate prediction and estimation of time-series. Due to their complexity, such naive extensions require high computational effort before acquiring acceptable behavior. Conditional RBMs (CRBMs) remedy this problem by proposing an alternative extension of RBMs~\cite{taylorcrbmicml}. Here, the architecture consists of two separate visible layers, representing history (i.e., values from previous time frames), and current values, and a hidden layer for latent correlation discovery. Though successful, CRBMs are only capable of modeling time series data with relatively ``smooth'' variations and similarly with other state-of-the-art neural network architectures for time series, e.g. recurrent neural networks, they can not learn within the same model different types of time-series. Thus, to model different types of non-linear time variations within the same model, the authors in~\cite{taylorcrbmicml} extend CRBMs by allowing for a three-way weight tensor connection among the different layers. Computational complexity is then reduced by adapting a factored version (i.e., FCRBMs) of the weight tensor, which leads to a construction exhibiting accurate modeling and prediction results in a variety of experiments, including human motion styles~\cite{gwtaylorhdts}. However, these methods fail to perform both classification and regression in one unified framework. Recently, Factored Four-Way Conditional Restricted Boltzmann Machines (FFW-CRBMs) have been proposed~\cite{ffwcrbmprl}. These extend FCRBMs by incorporating a label layer and a four-way weight tensor connection among the layers to  modulate the weights for capturing subtle temporal differences. This construction allowed FFW-CRBMs to perform both, i.e. classification and real-valued predictions, within the same model, and to outperform state-of-the-art specialized methods for classification or prediction~\cite{ffwcrbmprl}. 

\textbf{Contributions:} In this paper we, first, propose the use of FFW-CRBMs to estimate 3D trajectories from their 2D projections, while at the same time being also capable to classify those trajectories. Though successful, we discovered that FFW-CRBMs require substantial amount of \emph{labeled data} before achieving acceptable performance when predicting three-dimensional trajectories from two-dimensional projections. As FFW-CRBMs require \emph{three-dimensional labeled} information for accurate predictions which is not typically available, secondly, in this paper, we remedy these problems by proposing an extension of FFW-CRBMs, dubbed Disjunctive FFW-CRBMs (DFFW-CRBMs). Our extension refines the factoring of the four-way weight tensor connecting the machine layers to settings where labeled data is scarce. Adopting such a factorization ``specializes'' FFW-CRBMs and ensures lower energy levels  (approximately three times less energy on the overall dataset). This yields the sufficiency of a reduced training dataset for DFFW-CRBMs to reach similar classification performance to state-of-the-art methods and to at least double the performance on real-valued predictions. Importantly, such accuracy improvements come at the same computational cost of $\mathcal{O}\left(n^{2}\right)$ compared to FFW-CRBMs. Precisely, our machine requires limited labeled data (less than 10 $\%$ of the overall dataset) for: \textit{i)} simultaneously classifying and predicting three-dimensional trajectories based on their two-dimensional projections, and \textit{ii)} accurately estimating three-dimensional postures up to an arbitrary number of time-steps in the future. 

We have extensively tested DFFW-CRBMs on both, simulated and real-world data, to show that they are capable of outperforming state-of-the-art methods in real-valued predictions and classifications. In the first set of experiments we evaluate its performance by predicting and classifying simulated three-dimensional ball trajectories  (based on a real-world physics simulator) thrown from different initial spins. Given these successes, in the second set of experiments we predict and classify high-dimensional human poses and activities (up-to 32 human skeleton joints in 2D and 3D coordinates systems, corresponding to 160 dimensions) using real-world data showing that DFFW-CRBMs acquire double accuracy results at reduced labeled data sizes. 

\section{Background}
This section provides relevant background knowledge essential to the remainder of the paper. Firstly, restricted Boltzmann machines (RBMs), being at the basis of our proposed method, are surveyed. Secondly, Contrastive Divergence, a training algorithm for Deep Learning methods, is presented. The section concludes with a brief description of deep-learning based models for time series prediction and classification.
\label{sec:background}
\subsection{Restricted Boltzmann Machines}
Restricted Boltzmann machines (RBMs)~\cite{originalrbm} are energy-based models for unsupervised learning. They use a generative model of the distribution of training data for prediction~\cite{mocanugenerativereplay}. These models employ stochastic nodes and layers, making them less vulnerable to local minima~\cite{gwtaylorhdts}. Further, due to their stochastic neural configurations, RBMs possess excellent generalization and density estimation capabilities~\cite{bengiodl, mocanumljxbm}. 

Formally, an RBM consists of visible and hidden binary layers connected by an undirected bipartite graph. More exactly, the visible layer $\textbf{v}=[v_{1},\dots,v_{n_{v}}]$ collects all visible units $v_{i}$ and represents the real-data, while the hidden layer $\textbf{h}=[h_{1},\dots,h_{n_{h}}]$ representing all the hidden units $h_{j}$ increases the learning capability by enlarging the class of distributions that can be represented to an arbitrary complexity. $n_{v}$ and $n_{h}$ are the number of neurons in the visible and hidden layers, respectively. $W_{ij}$ denotes the weight connection between the $i^{th}$ visible and $j^{th}$ hidden unit, and $v_{i}$ and $h_{j}$ denote the state of the $i^{th}$ visible and $j^{th}$ hidden unit, respectively. The matrix of all weights between the layers is given by $\textbf{W}\in \mathbb{R}^{n_{h}\times n_{v}}$. The energy function of RBMs is given by
\begin{equation}
E\left(v,h\right)=-\sum_{i=1}^{n_{v}}\sum_{j=1}^{n_{h}}W_{ij}v_{i}h_{j}-\sum_{i=1}^{n_{v}}a_{i}v_{i} - \sum_{j=1}^{n_{h}}b_{j}h_{j}
\label{eq:rbmenergy}
\end{equation}
where, $a_{i}$ and $b_{j}$ represent the biases of the visible and hidden layers, respectively. The joint probability of a visible and hidden configuration can be written as $P\left(v,h\right)=\frac{\exp(-E(v,h))}{Z}$ with $Z=\sum_{x,y} \exp\left(-E(x,y)\right)$. The marginal distribution, $p(v)=\sum_{h}p(v,h)$ , can be used to determine the probability of a data point represented by a state $v$. 

\subsection{Training an RBM via Contrastive Divergence}
The RBMs parameters are trained by maximizing the likelihood function, typically by following the gradient of the energy function. Unfortunately, in RBMs, maximum likelihood estimation can not be applied directly due to intractability problems. These problems can be circumvented by using Contrastive Divergence (CD)~\cite{hintoncd} to train the RBM.
 In CD, learning follows the gradient of: 
\begin{equation}
CD_{n}\propto D_{KL}(p_{0}(\textbf{x})||p_{\infty}(\textbf{x}))-D_{KL}(p_{n}(\textbf{x})||p_{\infty}(\textbf{x}))
\end{equation}
where, $p_{n}(\cdot)$ is the distribution of a Markov chain running for $n$ steps and $D_{KL}$ symbolizes the Kullback-Leibler divergence~\cite{Ponti2017470}. To find the update rules for the free parameters of the RBM (i.e weights and biases), the RBM's energy function from  Equation~\ref{eq:rbmenergy} has to be differentiated with respect to those parameters. Thus, in $CD_{n}$ the weight updates are done as follows: 
$
w^{\tau+1}_{ij}=w^{\tau}_{ij}+\alpha\left(\left\langle\langle h_{j}v_{i}\rangle_{p(\textbf{h}|\textbf{v};\textbf{W})}\right\rangle_{0}-\langle h_{j}v_{i}\rangle_{n}\right)
$
where $\tau$ is the iteration number, $\alpha$ is the learning rate, 
$
\left\langle\langle h_{j}v_{i}\rangle_{p(\textbf{h}|\textbf{v};\textbf{W})}\right\rangle_{0}=\frac{1}{N_I}\sum_{k=1}^{N_I}v^{(q)}_{i}P(h^{(q)}_{j}=1|\textbf{v}^{(q)};\textbf{W})
$ and 
$
\langle h_{j}v_{i}\rangle_{n}=\frac{1}{N_I}\sum_{k=1}^{N_I}v^{(q)(n)}_{i}
P(h^{(q)(n)}_{j}=1|\textbf{v}^{(q)(n)};\textbf{W})
$
where $N_I$ is the total number of input instances, and the superscript $^{(q)}$ shows the $q^{th}$ input instance. The superscript $^{(n)}$ indicates that the states are obtained after $n$ steps of Gibbs sampling on a Markov chain which starts at the original data distribution $p_{0}(\cdot)$. In practice, learning can be performed using just one step Gibbs sampling, which is carried in four sub-steps: (1) initialize visible units, (2) infer all the hidden units, (3) infer all the visible units, and (4) update the weights and the biases.

\subsection{Factored Conditional Restricted Boltzmann Machine}\label{Sec:FCRBMs}
Conditional Restricted Boltzmann Machines (CRBM)~\cite{gwtaylorhdts} are an extension of RBMs used to model time series data, for example, human activities.  They use an undirected model with binary hidden variables connected to real-valued visible ones. At each time step $t$, the hidden and visible nodes receive a connection from the visible variables at the last $L$ time-steps. The history of the real-world values until time $t$ is collected in the real-valued history vector $\textbf{v}_{<t}$ with $n_{v_{<t}}=n_v(L-1)$ being the number of elements in $\textbf{v}_{<t}$. The total energy of the CRBM is given by: 
\begin{equation}
E=\sum_{i=1}^{n_v}\frac{(\hat{a}_{i,t}-v_{i,t})^{2}}{2\sigma^{2}_{i}}-\sum_{j=1}^{n_h}\hat{b}_{j,t}h_{j,t}-\sum_{i=1}^{n_v}\sum_{j=1}^{n_h}W_{ij}\frac{v_{i,t}}{\sigma_{i}}h_{j,t}
\end{equation}
where $\hat{a}_{i,t}=a_{i}+\sum_{k=1}^{n_{v_{<t}}}A_{ki}v_{k,<t}$ and $\hat{b}_{j,t}=b_{j}+\sum_{k=1}^{n_{v_{<t}}}B_{kj}v_{k,<t}$ represent the ``dynamic biases", with $k$ being the index of the elements from $\textbf{v}_{<t}$.  

Taylor and Hinton introduced the Factored Condition Restricted Boltzmann Machine (FCRBM)~\cite{gwtaylorhdts}, which permits the modeling of different styles of time series within the same model, due to the introduction of multiplicative, three-way interactions and of a \emph{preset} style label, $\mathbf{y}_{t}$. To reduce the computational complexity of this model, they factored the third order tensors between layer in products of matrices. Formally, FCRBM defines a joint probability distribution over the visible $\mathbf{v_{t}}$ and hidden $\mathbf{h_{t}}$ neurons. The joint distribution is conditioned on the past $L$ observations, $\mathbf{v_{<t}}$, model parameters, $\boldsymbol{\Theta}$, and the \emph{preset} style label, $\mathbf{y}_{t}$. Interested readers are referred to~\cite{gwtaylorhdts} for a more comprehensive discussion on CRBMs and FCRBMs. 

\subsection{Four-Way Conditional Restricted Boltzmann Machines}
Due to the limitations exhibited by FCRBMs, e.g., the impossibility of performing classification without extensions, we proposed the four-way conditional restricted Boltzmann machines (FW-CRBMs) for performing prediction and classification in one unified framework~\cite{ffwcrbmprl}. FW-CRBMs introduced an additional layer and a four-way multiplicative weight tensor interaction between neurons. Please note that, later on, other four-way models have been proposed but they can perform just classification an no prediction~\cite{Elaiwat2016152}. 

\begin{figure}[t]
\begin{center}
   \includegraphics[width=0.6\linewidth]{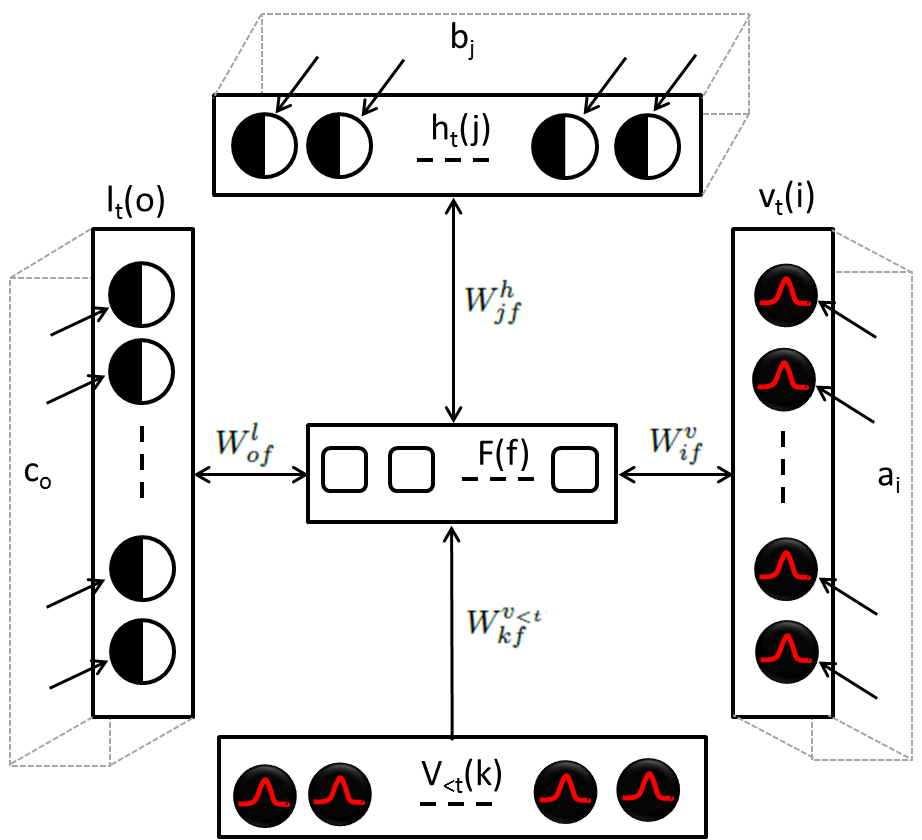}
\end{center}
\caption{A high level depiction of the FFW-CRBM showing the four layer configuration and the factored weight tensor connection among them. Gaussian nodes shown on the history and visible layers represent real-valued inputs, while sigmoidal nodes on the hidden and label layers demonstrate binary values.}
\label{fig:ffwcrbm}
\end{figure}

FW-CRBMs extended FCRBMs to include a label layer $\textbf{l}_{t}$ and a fourth order weight tensor connection $\mathbf{W}_{ijko} \in \mathbb{R}^{n_{\textbf{v}} \times n_{\textbf{h}} \times n_{\textbf{v}_{<t}} \times n_{\textbf{l}}}$, where $n_{\textbf{v}}$, $n_{\textbf{h}}$, $n_{\textbf{v}_{<t}}$, $n_{\textbf{l}}$ represent the number of neurons from the present, hidden, history and label layers, respectively. Though successful, FW-CRBMs exhibited high computational complexities (i.e., $\mathcal{O}\left(n^{4}\right)$) for tuning free parameters. Circumventing these problems, we factored the weight tensor into sums of products leading to more efficient machines (i.e., $\mathcal{O}\left(n^{2}\right)$) labeled as factored four-way conditional restricted Boltzmann machines (FFW-CRBMs). FFW-CRBMs, shown in Figure~\ref{fig:ffwcrbm}, minimize the following energy functional 
\begin{equation}
\label{eq:eqffwcrbm123}
\begin{split}
\mathbf{E}&(\textbf{v}_{t},\textbf{h}_{t},\textbf{l}_{t}|\textbf{v}_{<t},\Theta) =   \\
&\hspace{-0.5em}-\sum\limits_{i=1}^{n_v} \frac{{(v_{i,t}-a_i)}^2}{{\sigma_i}^2}-\sum\limits_{j=1}^{n_h} h_{j,t}b_j 
-\sum\limits_{o=1}^{n_l} l_{o,t}c_o \\
&\hspace{-0.5em}-\sum\limits_{f=1}^{n_F}\sum\limits_{i=1}^{n_v}W_{if}^{v}\frac{v_{i,t}}{\sigma_i}\sum\limits_{j=1}^{n_h} W_{jf}^{h}h_{j,t}\sum\limits_{k=1}^{n_{v<t}} W_{kf}^{v_{<t}}\frac{v_{k,<t}}{\sigma_k}\sum\limits_{o=1}^{n_l} W_{of}^{l}l_{o,t}
\end{split}
\end{equation}
where ${n_F}$ is number of factors and $i$, $j$, $k$, and $o$ are the indices of the visible layer neurons $\mathbf{v}_{t}$, the hidden layer neurons $\mathbf{h}_{t}$, the history layer neurons $\mathbf{v}_{<t}$ and the labeled layer neurons $\mathbf{l}_{t}$ respectively. $\mathbf{W}^v$, $\mathbf{W}^h$, $\mathbf{W}^l$ symbolize the bidirectional and symmetric weights from the visible, hidden and label layers to the factors, respectively, while $\mathbf{W}^{v_{<t}}$ represents the directed weights from the history layer to the factors. As in the case of the three-way models~\cite{modellingjointdensities}, standard CD is unsuccessful in training also the four-way models, due to the need of predicting two output layers (i.e. label and present layers). Thus, in~\cite{ffwcrbmprl} we proposed a sequential variant of CD, named sequential Markov chain contrastive divergence, more suitable for tuning the free parameters in FW-CRBMs.

FFW-CRBMs have shown good generalization and time series latent feature learning capabilities compared to state-of-the-art techniques including but not limited to, support vector machines, CRBMs, and FCRBMs~\cite{ffwcrbmprl}. It is for these reasons that we believe that FFW-CRBMs can serve as a basis for predicting three-dimensional trajectories from two-dimensional projections. Unfortunately, FFW-CRBMs are not readily applicable to such a problem as they require substantial amount of labeled data for successful tuning. In this paper, we extend FFW-CRBMs to Disjunctive FFW-CRBMs (DFFW-CRBMs) by proposing a novel factoring process essential for predicting and classifying 3D trajectories from 2D projections. Our model, detailed next, reduces sample complexities of current methods and allows for lower energy levels compared to FFW-CRBMs leading to improved performance.

\section{Disjunctive Factored Four Way Conditional restricted Boltzmann Machines}
\label{sec:newmodel}
\begin{figure}[t]
\begin{center}
   \includegraphics[width=0.6\linewidth]{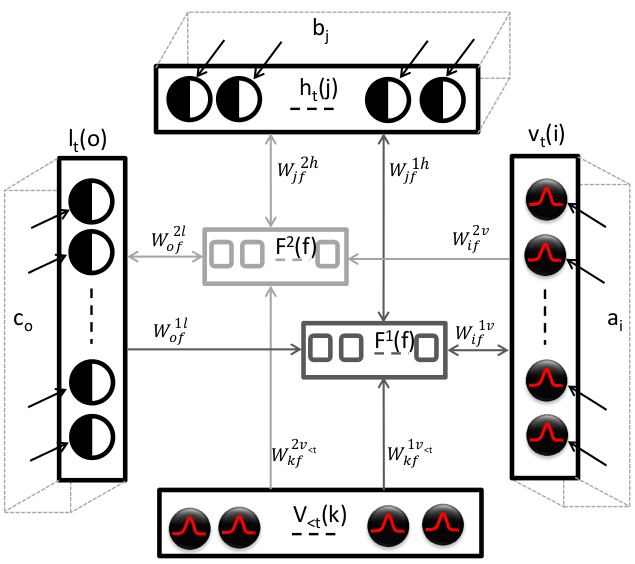}
\end{center}
\caption{A high level depiction of DFFW-CRBMs showing the four layer configuration and the refined tensors factoring for increased accuracy and efficiency.}
\label{fig:dffwcrbm}
\end{figure}
This section details disjunctive factored four way conditional restricted Boltzmann machines (DFFW-CRBMs), shown in Figure~\ref{fig:dffwcrbm}. Similarly to FFW-CRBMs, our model consists of four layers to represent visible, history, hidden, and label units. Contrary to the factoring adopted by FFW-CRBMs, however, our model incorporates two new factoring layers. The first, i.e., $F^{1}(f)$ in the figure, is responsible for specializing the machine to real-valued predictions through $\textbf{W}^{1l}$, $\textbf{W}^{1v}$, $\textbf{W}^{1h}$, and $\textbf{W}^{1v_{<t}}$, while the second, $F^{2}(f)$, specializes the machine to classification through the corresponding weight tensor collections. Such a specialization is responsible for reducing sample complexities needed by DFFW-CRBMs for successful parameter tuning as demonstrated in Section~\ref{sec:experiments}, while the computational complexity of DFFW-CRBM remains the same as for FFW-CRBM (i.e., $\mathcal{O}\left(n^{2}\right)$) . Given our novel construction, DFFW-CRBMs require their own special mathematical treatment. Next, we detail each of the energy functional and learning rules needed by DFFW-CRBMs.

\subsection{DFFW-CRBM's Energy Function}
The energy function of DFFW-CRBMs consists of three major terms. The first, i.e., $-\sum\limits_{i=1}^{n_v} \frac{{(v_{i,t}-a_i)}^2}{{\sigma_i}^2}-\sum\limits_{j=1}^{n_h} h_{j,t}b_j 
-\sum\limits_{o=1}^{n_l} l_{o,t}c_o$ corresponds to the standard energy representing a specific submachine of DFFW-CRBMs (i.e. the energy given by the neurons of each layers and their biases) while the second two denote energies related to the first and second factoring layers, respectively: 
\begin{equation}
\begin{split}
\mathbf{E}&(\textbf{v}_{t},\textbf{h}_{t},\textbf{l}_{t}|\textbf{v}_{<t},\Theta) =  \label{Eq1:EnergyDFFWCRBM}\\
&\hspace{-0.5em}-\underbrace{\sum\limits_{i=1}^{n_v} \frac{{(v_{i,t}-a_i)}^2}{{\sigma_i}^2}-\sum\limits_{j=1}^{n_h} h_{j,t}b_j 
-\sum\limits_{o=1}^{n_l} l_{o,t}c_o}_{\text{standard three-layer energy}} \\
&\hspace{-0.5em}-\underbrace{\sum\limits_{f=1}^{n_{F^1}}\sum\limits_{i=1}^{n_v}W_{if}^{1v}\frac{v_{i,t}}{\sigma_i}\sum\limits_{j=1}^{n_h} W_{jf}^{1h}h_{j,t}\sum\limits_{k=1}^{n_{v<t}} W_{kf}^{1v_{<t}}\frac{v_{k,<t}}{\sigma_k}\sum\limits_{o=1}^{n_l} W_{of}^{1l}l_{o,t}}_{\text{first factoring layer}}  \\ 
&\hspace{-0.5em}-\underbrace{\sum\limits_{f=1}^{n_{F^2}}\sum\limits_{i=1}^{n_v}W_{if}^{2v}\frac{v_{i,t}}{\sigma_i}\sum\limits_{j=1}^{n_h} W_{jf}^{2h}h_{j,t}\sum\limits_{k=1}^{n_{v<t}} W_{kf}^{2v_{<t}}\frac{v_{k,<t}}{\sigma_k}\sum\limits_{o=1}^{n_l} W_{of}^{2l}l_{o,t}}_{\text{second factoring layer}}  
\end{split}
\end{equation} 
Here, $n_{F^1}$ denotes the total number of factors for the weight tensor collection specializing DFFW-CRBMs to regression, while $n_{F^2}$ is total the number of factors responsible for classification. $i$, $j$, $k$, and $o$ represent the indices of the visible layer neurons $\mathbf{v}_{t}$, the hidden layer neurons $\mathbf{h}_{t}$, the history layer neurons $\mathbf{v}_{<t}$ and the labeled layer neurons $\mathbf{l}_{t}$, respectively. Furthermore, $\mathbf{W}^{1v}$ and $\mathbf{W}^{1h}$ represent the bidirectional and symmetric weight connections from the visible and hidden layers to the factors,  while $\mathbf{W}^{1l}$ and $\mathbf{W}^{1v_{<t}}$ denote the \emph{directed} weights from the label and history layers to the factors. Similarly, $\mathbf{W}^{2l}$ and $\mathbf{W}^{2h}$ represent the bidirectional and symmetric weights from the label and hidden layers to the factors,  while $\mathbf{W}^{2v}$ and $\mathbf{W}^{2v_{<t}}$ denote the directed weights from the visible and history layers to the factors. Finally, the two groups of four weight matrices each noted with $\mathbf{W}^{1.}$ and $\mathbf{W}^{2.}$ belong to the factorized tensor specialization in regression and classification, respectively.

\subsection{DFFW-CRBM's Activation Probabilities}
Inference for DFFW-CRBM corresponds to determining values of the activation probabilities for each of the units. 
As shown in Figure~\ref{fig:dffwcrbm}, units within the same layer do not share connections. This allows for parallel probability computation for all units within the same layer. The overall input of each hidden $s^{h}_{j,t}$, visible $s^{v}_{i,t}$, and labelled $s^{l}_{o,t}$ unit is given by:  
\begin{equation}
\begin{split}
{s}_{j,t}^{h}&=\sum\limits_{f=1}^{n_{F^1}}W_{jf}^{1h}\sum\limits_{i=1}^{n_v}W_{if}^{1v}\frac{v_{i,t}}{\sigma_i}\sum\limits_{k=1}^{n_{n_{v<t}}} W_{kf}^{1v_{<t}}\frac{v_{k,<t}}{\sigma_k}\sum\limits_{o=1}^{n_l} W_{of}^{1l}l_{o,t}  \\ 
&+\sum\limits_{f=1}^{n_{F^2}}W_{jf}^{1h}\sum\limits_{i=1}^{n_v}W_{if}^{2v}\frac{v_{i,t}}{\sigma_i}\sum\limits_{k=1}^{n_{n_{v<t}}} W_{kf}^{2v_{<t}}\frac{v_{k,<t}}{\sigma_k}\sum\limits_{o=1}^{n_l} W_{of}^{2l}l_{o,t} \\ 
{s}_{i,t}^{v}&=\sum\limits_{f=1}^{n_{F^1}}W_{if}^{1v}\sum\limits_{j=1}^{n_h}W_{jf}^{1h}h_{j,t}\sum\limits_{k=1}^{n_{n_{v<t}}} W_{kf}^{1v_{<t}}\frac{v_{k,<t}}{\sigma_k}\sum\limits_{o=1}^{n_l} W_{of}^{1l}l_{o,t}
\\
{s}_{o,t}^{l}&=\sum\limits_{f=1}^{n_{F^2}}W_{of}^{2l}\sum\limits_{j=1}^{n_h} W_{jf}^{2h}h_{j,t}\sum\limits_{k=1}^{n_{n_{v<t}}} W_{kf}^{2v_{<t}}\frac{v_{k,<t}}{\sigma_k}\sum\limits_{i=1}^{n_v}W_{if}^{2v}\frac{v_{i,t}}{\sigma_i} 
\label{Eq:inputDFFWCRBM}
\end{split}
\end{equation}
Consequently, for each of the $j^{th}$ hidden, $i^{th}$ visible, and $o^{th}$ labelled units, the activation probabilities can be determined as
\begin{equation}
\begin{split}
p(h_{j,t}=1|\mathbf{v}_t,\mathbf{v}_{<t},\mathbf{l}_t)&=\frac{1}{1+e^{-\left(b_j+{s}_{j,t}^{h}\right)}} \\ 
p(v_{i,t}=x|\mathbf{h}_t,\mathbf{v}_{<t},\mathbf{l}_t)&=\mathcal{N}\left(a_i+ {s}_{i,t}^{v},\sigma_i^2\right) \\  
p(l_{o,t}=1|\mathbf{v}_t,\mathbf{v}_{<t},\mathbf{h}_t)&=\frac{1}{1+e^{-\left(c_o+{s}_{o,t}^{l}\right)}} 
\label{Eq:Eq:inferenceDFFWCRBM}
\end{split}
\end{equation}
where $\mathcal{N}(\cdot)$ represents the standard Gaussian distribution. 

\subsection{Parameter Tuning: Update Rules \& Algorithm}
\subsubsection{Update Rules}
Generally, parameters, $\mathbf{\Theta}$, are updated according to:
\begin{equation}
\mathbf{\Theta}_{\tau+1}=\mathbf{\Theta}_{\tau}+\underbrace{\rho\mathbf{\widetilde\Theta}_{\tau}+\alpha(\Delta\mathbf{\Theta}_{\tau+1}-\gamma\mathbf{\Theta}_{\tau})}_{\mathbf{\widetilde\Theta}_{\tau+1} \text{ update}}
\end{equation}
where $\tau$ represents the update iteration, $\rho\in(0,1)$ is the momentum, $\alpha\in(0,1)$ denotes the learning rate, and $\gamma\in(0,1)$ is the weight decay. A more detailed discussion on the choice of these parameters is provided by Hinton in~\cite{hintontrain}. Therein, the update rules are attained by deriving the energy functional with respect to free parameters (i.e., weights matrices, and the biases of each of the layers). In DFFW-CRBMs, a set of eight free parameters, corresponding to the connections between the factors and each of the layers, has to be inferred. These are presented below. Intuitively, each of these update equations, aims at minimizing the reconstruction error (i.e., the error between the original inputs and these reconstructed through the model). Moreover, each of the update equations include three main terms representing the connections between the factored weights and the corresponding layer of the machine, as per Figure~\ref{fig:dffwcrbm}. For instance, connections to only the hidden, history, and label layers suffice for updating $W^{1v}_{if}$. 
Thus, the update rules $\Delta\mathbf{\Theta}_{\tau}$ for each of the weights corresponding to the first factored layer, can be computed as:   
\begin{equation}
\begin{split}
{\Delta}W_{if}^{1v}&\propto{{\Bigg\langle}\displaystyle v_{i,t}\sum\limits_{j=1}^{n_h} W_{jf}^{1h}h_{j,t}\sum\limits_{k=1}^{n_{v<t}} W_{kf}^{1v_{<t}}v_{k,<t}\sum\limits_{o=1}^{n_l} W_{of}^{1l}l_{o,t}{\Bigg\rangle}}_{0}\\   
&\hspace{0em}-{{\Bigg\langle}\displaystyle v_{i,t}\sum\limits_{j=1}^{n_h} W_{jf}^{1h}h_{j,t}\sum\limits_{k=1}^{n_{v<t}} W_{kf}^{1v_{<t}}v_{k,<t}\sum\limits_{o=1}^{n_l} W_{of}^{1l}l_{o,t}{\Bigg\rangle}}_{\lambda} \\ 
{\Delta}W_{kf}^{1v_{<t}}&\propto{{\Bigg\langle}\displaystyle v_{k,<t}\sum\limits_{j=1}^{n_h} W_{jf}^{1h}h_{j,t}\sum\limits_{i=1}^{n_v} W_{if}^{1v}v_{i,t}\sum\limits_{o=1}^{n_l} W_{of}^{1l}l_{o,t}{\Bigg\rangle}}_{0}\\&\hspace{0em}-{{\Bigg\langle}\displaystyle v_{k,<t}\sum\limits_{j=1}^{n_h} W_{jf}^{1h}h_{j,t}\sum\limits_{i=1}^{n_v} W_{if}^{1v}v_{i,t}\sum\limits_{o=1}^{n_l} W_{of}^{1l}l_{o,t}{\Bigg\rangle}}_{\lambda} \\ 
{\Delta}W_{of}^{1l}&\propto{{\Bigg\langle}l_{o,t}\sum\limits_{k=1}^{n_{v<t}}W_{kf}^{1v_{<t}}v_{k,<t}\sum\limits_{j=1}^{n_h}W_{jf}^{1h}h_{j,t}\sum\limits_{i=1}^{n_v}W_{if}^{1v}v_{i,t}
{\Bigg\rangle}}_{0}\\&\hspace{0em}-{{\Bigg\langle}l_{o,t}\sum\limits_{k=1}^{n_{v<t}}W_{kf}^{1v_{<t}}v_{k,<t}\sum\limits_{j=1}^{n_h}W_{jf}^{1h}h_{j,t}\sum\limits_{i=1}^{n_v}W_{if}^{1v}v_{i,t}{\Bigg\rangle}}_{\lambda} \\ 
{\Delta}W_{jf}^{1h}&\propto{{\Bigg\langle}h_{j,t}\sum\limits_{k=1}^{n_{v<t}}W_{kf}^{1v_{<t}}v_{k,<t}\sum\limits_{i=1}^{n_v}W_{if}^{1v}v_{i,t}\sum\limits_{o=1}^{n_l}W_{of}^{1l}l_{o,t}
{\Bigg\rangle}}_{0}\\&\hspace{0em}-{{\Bigg\langle}h_{j,t}\sum\limits_{k=1}^{n_{v<t}}W_{kf}^{1v_{<t}}v_{k,<t}\sum\limits_{i=1}^{n_v}W_{if}^{1v}v_{i,t}\sum\limits_{o=1}^{n_l}W_{of}^{1l}l_{o,t}{\Bigg\rangle}}_{\lambda}  
\end{split}
\end{equation}
while for the second factoring we have: 
\begin{equation}
\begin{split}
{\Delta}W_{if}^{2v}&\propto{{\Bigg\langle}\displaystyle v_{i,t}\sum\limits_{j=1}^{n_h} W_{jf}^{2h}h_{j,t}\sum\limits_{k=1}^{n_{v<t}} W_{kf}^{2v_{<t}}v_{k,<t}\sum\limits_{o=1}^{n_l} W_{of}^{2l}l_{o,t}{\Bigg\rangle}}_{0}\\ &\hspace{0em}-{{\Bigg\langle}\displaystyle v_{i,t}\sum\limits_{j=1}^{n_h} W_{jf}^{2h}h_{j,t}\sum\limits_{k=1}^{n_{v<t}} W_{kf}^{2v_{<t}}v_{k,<t}\sum\limits_{o=1}^{n_l} W_{of}^{2l}l_{o,t}{\Bigg\rangle}}_{\lambda} \\
{\Delta}W_{kf}^{2v_{<t}}&\propto{{\Bigg\langle}\displaystyle v_{k,<t}\sum\limits_{j=1}^{n_h} W_{jf}^{2h}h_{j,t}\sum\limits_{i=1}^{n_v} W_{if}^{2v}v_{i,t}\sum\limits_{o=1}^{n_l} W_{of}^{2l}l_{o,t}{\Bigg\rangle}}_{0}\\&\hspace{0em}-{{\Bigg\langle}\displaystyle v_{k,<t}\sum\limits_{j=1}^{n_h} W_{jf}^{2h}h_{j,t}\sum\limits_{i=1}^{n_v} W_{if}^{2v}v_{i,t}\sum\limits_{o=1}^{n_l} W_{of}^{2l}l_{o,t}{\Bigg\rangle}}_{\lambda} \\
{\Delta}W_{of}^{2l}&\propto{{\Bigg\langle}l_{o,t}\sum\limits_{k=1}^{n_{v<t}}W_{kf}^{2v_{<t}}v_{k,<t}\sum\limits_{j=1}^{n_h}W_{jf}^{2h}h_{j,t}\sum\limits_{i=1}^{n_v}W_{if}^{2v}v_{i,t}
{\Bigg\rangle}}_{0}\\&\hspace{0em}-{{\Bigg\langle}l_{o,t}\sum\limits_{k=1}^{n_{v<t}}W_{kf}^{2v_{<t}}v_{k,<t}\sum\limits_{j=1}^{n_h}W_{jf}^{2h}h_{j,t}\sum\limits_{i=1}^{n_v}W_{if}^{2v}v_{i,t}{\Bigg\rangle}}_{\lambda} \\
{\Delta}W_{jf}^{2h}&\propto{{\Bigg\langle}h_{j,t}\sum\limits_{k=1}^{n_{v<t}}W_{kf}^{2v_{<t}}v_{k,<t}\sum\limits_{i=1}^{n_v}W_{if}^{2v}v_{i,t}\sum\limits_{o=1}^{n_l}W_{of}^{2l}l_{o,t}
{\Bigg\rangle}}_{0}\\&\hspace{0em}-{{\Bigg\langle}h_{j,t}\sum\limits_{k=1}^{n_{v<t}}W_{kf}^{2v_{<t}}v_{k,<t}\sum\limits_{i=1}^{n_v}W_{if}^{2v}v_{i,t}\sum\limits_{o=1}^{n_l}W_{of}^{2l}l_{o,t}{\Bigg\rangle}}_{\lambda}  
\end{split}
\end{equation}
and for the biases are: 
\begin{equation}
\begin{split}
{\Delta}a_{i}&\propto{{\langle}v_{i,t}{\rangle}}_{0}-{{\langle}v_{i,t}{\rangle}}_{\lambda}
\\
 {\Delta}b_{j}&\propto{{\langle}h_{j,t}{\rangle}}_{0}-{{\langle}h_{j,t}{\rangle}}_{\lambda}
\\
{\Delta}c_{o}&\propto{{\langle}l_{o,t}{\rangle}}_{0}-{{\langle}l_{o,t}{\rangle}}_{\lambda}
\end{split}
\end{equation}
where $\lambda$ represents a Markov chain step running for a total of $n$ steps and starting at the original data distribution, $\langle \cdot \rangle_{0}$ denotes the expectation under the input data, and $\langle \cdot \rangle_{\lambda}$ represents the model's expectation. 

\subsubsection{Sequential CD for DFFW-CRBMs}
Algorithm~\ref{Algo:AlgoOne} presents a high-level description of the sequential Markov chain contrastive divergence~\cite{ffwcrbmprl} adapted to train DFFW-CRBMs. It shows the two main steps needed for training such machines. Firstly, the visible layer is inferred by fixing the history and label layers. While in the second step the label layer is reconstructed  by fixing the history and the present layers. Updating the weights involves the implementation of the rules derived in the previous section. 
These two procedures are then repeated for a pre-specified number of epochs, where at each epoch the reconstruction error is decreasing to reach the minimum of the energy function, guaranteeing a minimized divergence between the original data distribution and the one given  by the model.

\begin{algorithm}[ht!]
\footnotesize
\caption{Sequential Contrastive Divergence for DFFW-CRBMs}
\hrule 
\hrule
\textbf{Inputs:} TD - training data, $n$ - number of Markov Chain steps\;
 \hrule
\textbf{Initialization:} $\mathbf{\Theta}$ $\leftarrow$ $\mathcal{N}(0,\sigma^2)$, Set $\alpha$, $\rho$, $\gamma$\;
 \hrule
 \For{all epochs}{
  \For{each Sample $\in$ TD}{
    \%\%\textit{First Markov Chain to reconstruct $\mathbf{v}_t$}\;
    Init $\mathbf{v}_t$ $\leftarrow$ 0, $\mathbf{l}_t$=Sample.Label, $\mathbf{v}_{<t}$=Sample.History\;
    $\mathbf{h}_{t}$ = InferHiddenLayer($\mathbf{v}_t$,$\mathbf{l}_t$,$\mathbf{v}_{<t}$,$\mathbf{\Theta}$)\;
    \For{$\lambda=0$;$\lambda<n$;$\lambda++$}{
      \%\%\textit{Positive phase}\;
      \textbf{pSt}=GetPosStats($\mathbf{h}_t$,Sample.Present,$\mathbf{l}_t$,$\mathbf{v}_{<t}$,$\mathbf{\Theta}$)\;
      \%\%\textit{Negative phase}\;      
      $\mathbf{v}_t$=InferPresentLayer($\mathbf{h}_t$,$\mathbf{l}_t$,$\mathbf{v}_{<t}$,$\mathbf{\Theta}$)\;
      $\mathbf{h}_{t}$ = InferHiddenLayer($\mathbf{v}_t$,$\mathbf{l}_t$,$\mathbf{v}_{<t}$,$\mathbf{\Theta}$)\;
      \textbf{nSt}=GetNegStats($\mathbf{h}_t$,$\mathbf{v}_t$,$\mathbf{l}_t$,$\mathbf{v}_{<t}$,$\mathbf{\Theta}$)\;
      $\mathbf{\Theta}$=UpdateWeights(\textbf{pSt},\textbf{nSt},$\mathbf{\Theta}$,$\alpha$,$\rho$,$\gamma$)\;
    }
    \%\%\textit{Second Markov Chain to reconstruct $\mathbf{l}_t$}\;
    Init $\mathbf{l}_t$ $\leftarrow$ 0, $\mathbf{v}_t$ = Sample.Present, $\mathbf{v}_{<t}$ = Sample.History\;
    $\mathbf{h}_{t}$ = InferHiddenLayer($\mathbf{v}_t$,$\mathbf{l}_t$,$\mathbf{v}_{<t}$,$\mathbf{\Theta}$)\;
    \For{$\lambda=0$;$\lambda<n$;$\lambda++$}{
      \%\%\textit{Positive phase}\;
      \textbf{pSt}=GetPosStats($\mathbf{h}_t$,Sample.Label,$\mathbf{v}_t$,$\mathbf{v}_{<t}$,$\mathbf{\Theta}$)\;
      \%\%\textit{Negative phase}\;      
      $\mathbf{l}_t$=InferLabelLayer($\mathbf{h}_t$,$\mathbf{v}_t$,$\mathbf{v}_{<t}$,$\mathbf{\Theta}$)\;
      $\mathbf{h}_{t}$ = InferHiddenLayer($\mathbf{v}_t$,$\mathbf{l}_t$,$\mathbf{v}_{<t}$,$\mathbf{\Theta}$)\;
      \textbf{nSt}=GetNegStats($\mathbf{h}_t$,$\mathbf{v}_t$,$\mathbf{l}_t$,$\mathbf{v}_{<t}$,$\mathbf{\Theta}$)\;
      $\mathbf{\Theta}$=UpdateWeights(\textbf{pSt},\textbf{nSt},$\mathbf{\Theta}$,$\alpha$,$\rho$,$\gamma$)\;
    }    
  }
 }
 \hrule
\label{Algo:AlgoOne}
\end{algorithm}

\section{Experiments and Results}
\label{sec:experiments}
This section extensively tests the performance of DFFW-CRBMs on both simulated as well as on real-world datasets.  The major goal of these experiments was to assess the capability of DFFW-CRBM to predict three-dimensional trajectories from two-dimensional projection, given small amounts of labeled data (i.e., in the order of 9-10 \% of the total dataset). As a secondary objective, the goal was to classify such trajectories to different spins (ball trajectories) or activities (human pose estimation). In the real-valued prediction setting, we compared our method to state-of-the-art FFW-CRBMs and FCRBMs, while for classification our method's performance was tested against FFW-CRBMs and support vector machines with radial basis functions (SVM-RBFs)~\cite{vapniksvm}. 

\textbf{Evaluation Metrics:} To assess the models' performance, a variety of standard metrics were used. For classification, we used accuracy~\cite{roc} in percentages, while for estimation tasks, we used the Normalized Root Mean Square Error (NRMSE) estimating distance between the prediction and ground truth, Pearson Correlation Coefficient (PCC) reflecting the correlations between predictions and ground truth, and the P-value to arrive at statistically significant predictions.

\subsection{Ball Trajectory Experiments}

We generated different ball trajectories thrown with different spins using the Bullet Physics Library\footnote{http://bulletphysics.org, Last accessed on November $8^{th}$ 2016}. With this simulated dataset we targeted three objectives using small amounts (9 \%) of labeled training data. First, we estimated 3D ball coordinates based on their 2D projections at each time-step $t$ (i.e., one-step prediction). Second, we aimed at predicting near-future (i.e., couple of time steps in the future) 3D ball coordinates recursively, while giving limited 2D sequence of coordinates as a starting point. Third, we classified various ball spins based on just 2D coordinates. We used four trajectory classes corresponding to four different ball spin types. For each class, a set of 11 trajectories each containing approximately 400 time-steps (amounting to a total of 17211 data instances) were sampled. To assess the performance of DFFW-CRBM, we performed 11-fold cross validation and reported mean and standard deviation results. Precisely, from each class of trajectories we used only \emph{one labeled trajectory}\footnote{A labeled trajectory has complete information: the 3D ball coordinates, their 2D projections, and the spin (i.e. class).} to train the models and the other 10 were used for testing. 

\begin{figure}[t]
\begin{center}
   \includegraphics[width=0.6\linewidth]{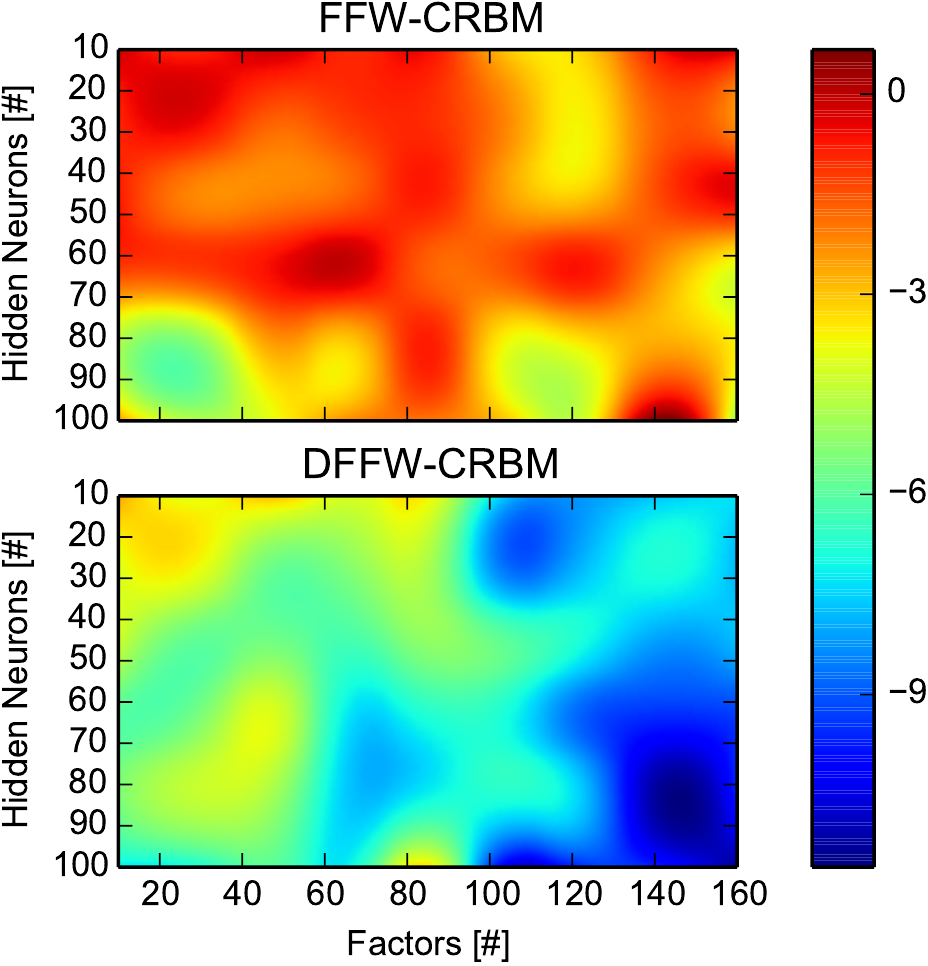}
\end{center}
\caption{Averaged energy levels of FFW-CRBM and DFFW-CRBM over all ball trajectories when the parameters (i.e. number of hidden neurons and factors) are varying. The training was done for 100 epochs.}
\label{fig:enleveltuneparameterballs}
\end{figure}
\begin{table}
\scriptsize
\tabcolsep=0.03cm
\begin{center}
\begin{tabular}{|c|c|c|c|r|r|r|}
\hline
\multicolumn{2}{|c|}{\multirow{2}{*}{Task}}&\multirow{2}{*}{Metrics}&\multicolumn{4}{c|}{Methods}\\
\cline{4-7}
\multicolumn{2}{|c|}{}&&SVM-RBF&FCRBM&FFW-CRBM&DFFW-CRBM \\
\hline\hline
\multicolumn{2}{|c|}{Classification}&Accuracy[\%]&39.26$\pm$4.63&N/A&37.49$\pm$3.66&\textbf{39.51$\pm$4.47}\\
\hline\hline
\multicolumn{2}{|c|}{Present Step}&NRMSE[\%]&N/A&18.38$\pm$8.07&19.53$\pm$33.24&\textbf{11.24$\pm$8.53}\\
\multicolumn{2}{|c|}{3D estimation}&PCC&N/A&-0.06$\pm$0.70&0.31$\pm$0.72&\textbf{0.62$\pm$0.61}\\
\multicolumn{2}{|c|}{}&P-value&N/A&0.51$\pm$0.29&0.40$\pm$0.29&\textbf{0.28$\pm$0.27}\\
\hline\hline
&After&NRMSE[\%]&N/A&25.61$\pm$3.25&23.53$\pm$2.48&\textbf{9.52$\pm$6.12}\\
&1 step&PCC&N/A&0.14$\pm$0.69&0.31$\pm$0.74&\textbf{0.95$\pm$0.14}\\
Multi-Step&&P-value&N/A&0.50$\pm$0.29&0.38$\pm$0.25&\textbf{0.12$\pm$0.18}\\
\cline{2-7}
3D&After&NRMSE[\%]&N/A&31.38$\pm$7.99&29.49$\pm$8.14&\textbf{19.93$\pm$10.27}\\
prediction&50 steps&PCC&N/A&0.05$\pm$0.69&-0.05$\pm$0.72&\textbf{0.20$\pm$0.66}\\
&&P-value&N/A&0.51$\pm$0.26&0.47$\pm$0.26&0.51$\pm$0.26\\
\hline
\end{tabular}
\end{center}
\caption{Classification, present step 3D estimation, and multi-step 3D prediction for the balls trajectories experiment. Results, cross-validated and presented with mean and standard deviation, show that our method is capable of outperforming state-of-the-art techniques on all evaluation metrics.}
\label{tab:claspreddiffballs}
\end{table}

\textbf{Deep Learner Setting:} The visible layers of both models (i.e. FFW-CRBM and DFFW-CRBM) were set to 5 neurons, three denoting 3D ball center coordinates (i.e. x, y, z), and two for its 2D projection at time $t$. The label layer consisted of 4 neurons (one for each of the different spins classes), while the history layers included 100 neurons corresponding to the last 50 history frames. One frame incorporates the 2D coordinates of the center of the ball projected in a two dimensional space. The number of hidden neurons was set to $10$, and the number of factors to $100$, as discussed in the next paragraph, and in Subsection~\ref{subsec:har}. A learning rate of $10^{-4}$ and momentum of $0.5$ were chosen. Weight decay factors were set to $0.0002$, and the number of the Markov chain steps for CD in the training phase, but also for the Gibbs sampling in the testing phase, was set to 3. All weights were initialized with $\mathcal{N}(0,0.3)$. Finally, data were normalized to have 0 mean and unit variance as explained in~\cite{hintontrain}, and the models were trained for 100 epochs.

\textbf{Importance of disjunctive Factoring:} To find the optimal number of hidden neurons and factors, we have performed exhaustive search by varying the number of hidden neurons from 10 to 100 and the number of factors from 10 to 160.  To gain some insights on the behavioral differences between FFW-CRBMs and DFFW-CRBMs, even if the energy equation of DFFW-CRBM has an extra tensor, in Figure~\ref{fig:enleveltuneparameterballs} we illustrate  on the same scale the heat-map of the averaged energy levels. They were computed using Equation~\ref{eq:eqffwcrbm123} for FFW-CRBM and Equation~\ref{Eq1:EnergyDFFWCRBM} for DFFW-CRBM, after both models were trained for 100 epochs. Though both models acquire the lowest energy levels in a configuration starting with 10-20 hidden neurons and a number of factors larger than 100, analyzing these results signifies the importance of the disjunctive factoring introduced in the paper. Namely, DFFW-CRBMs always acquire lower energy levels compared to FFW-CRBMs due to it's ``specialized'' tensor factoring. Moreover, by averaging the energy levels from the aforementioned figure, we found that the average energy level of DFFW-CRBM is approximately three times smaller than the one of FFW-CRBM (i.e. $-6.63\pm2.09$ for DFFW-CRBM, and $-2.04\pm1.43$ for FFW-CRBM), thus anticipating the more accurate performance results, as showed next. 

\begin{figure*}[t]
\begin{center}
   \includegraphics[width=0.19\linewidth]{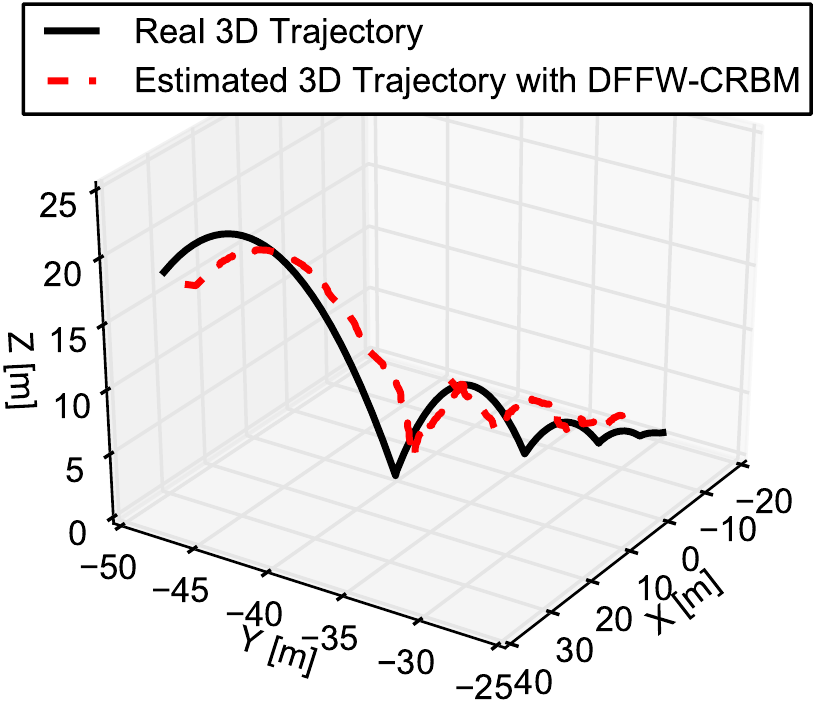}
   \includegraphics[width=0.19\linewidth]{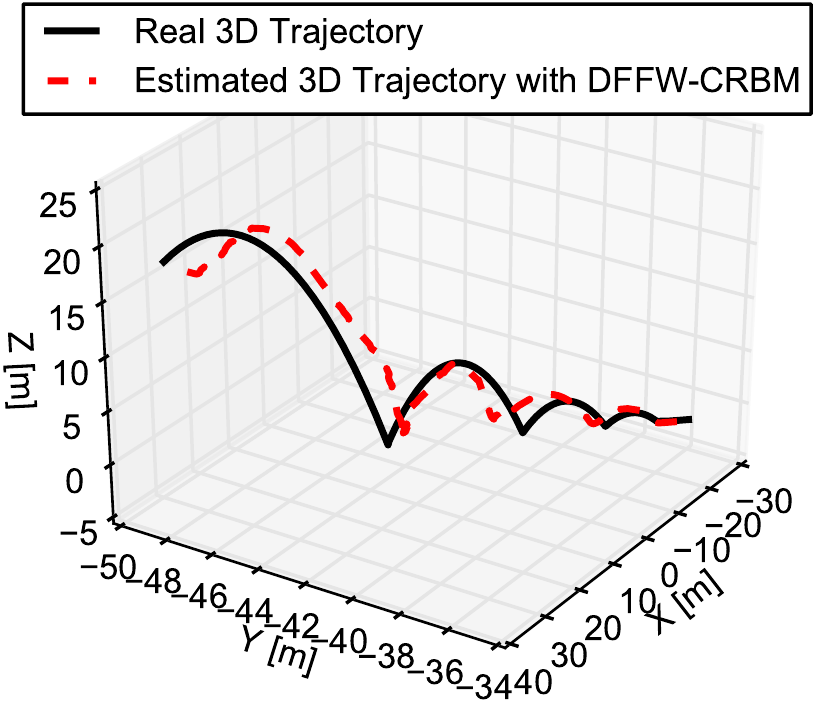}
   \includegraphics[width=0.19\linewidth]{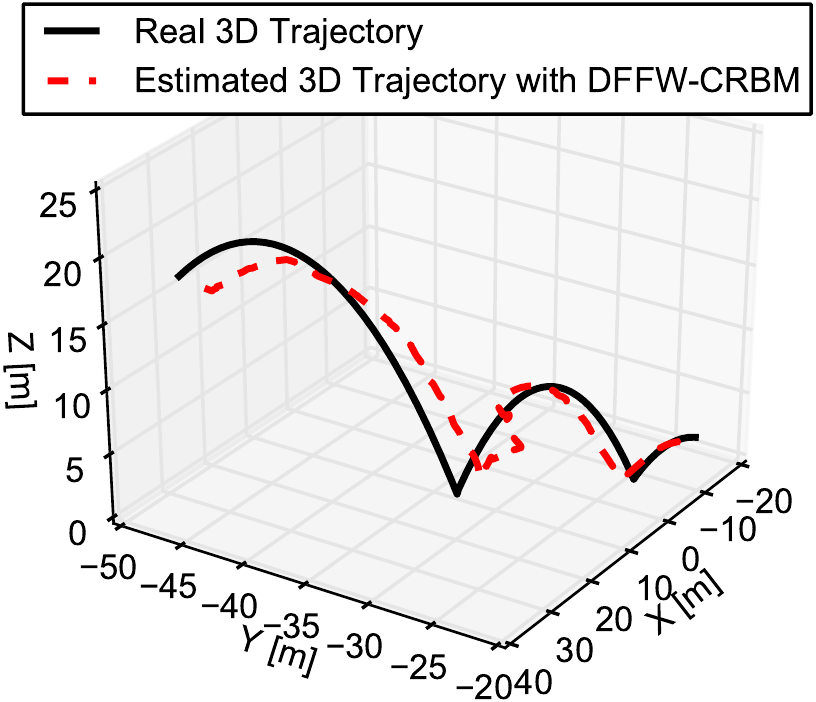}
   \includegraphics[width=0.19\linewidth]{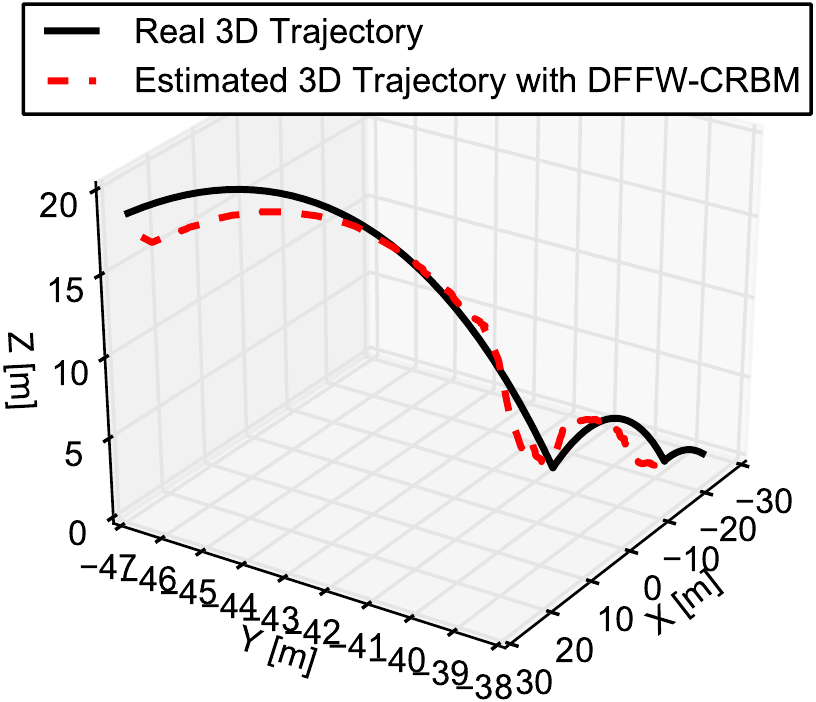}
   \includegraphics[width=0.19\linewidth]{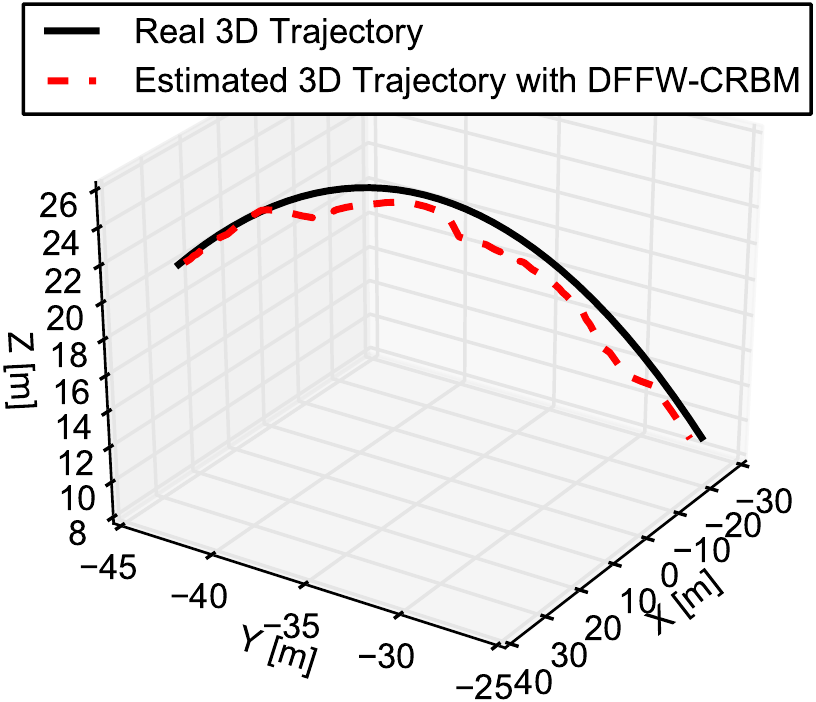}
   \includegraphics[width=0.19\linewidth]{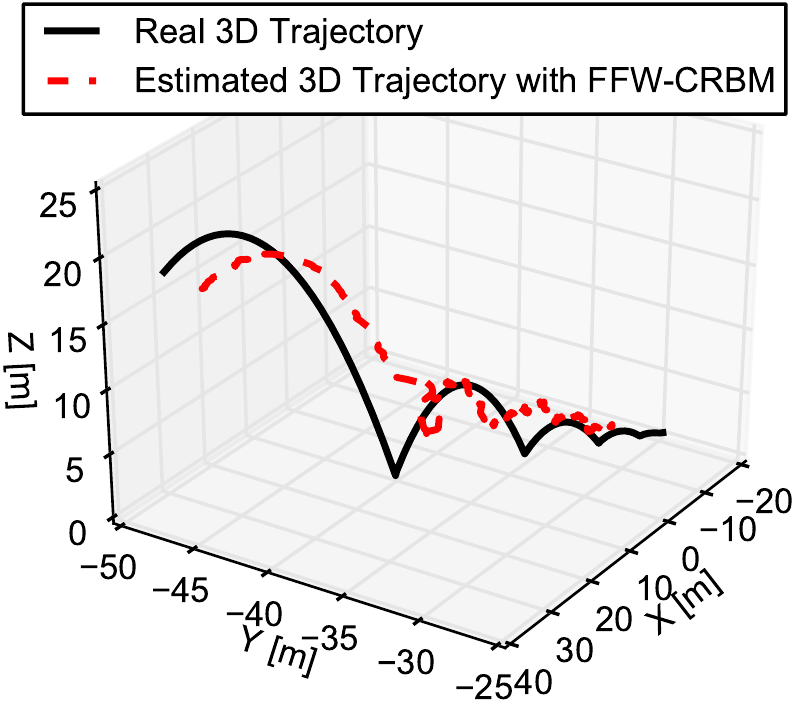}
   \includegraphics[width=0.19\linewidth]{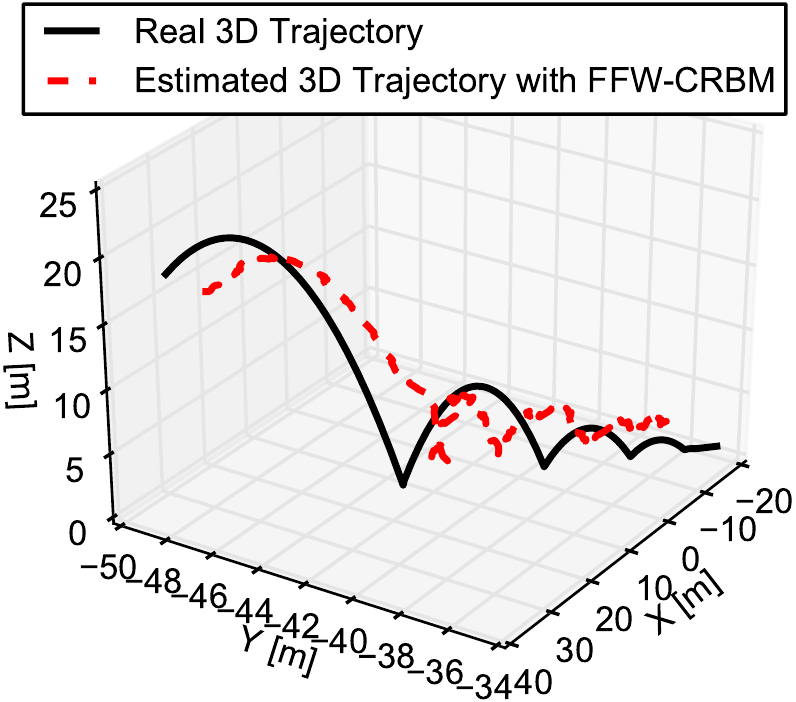}
   \includegraphics[width=0.19\linewidth]{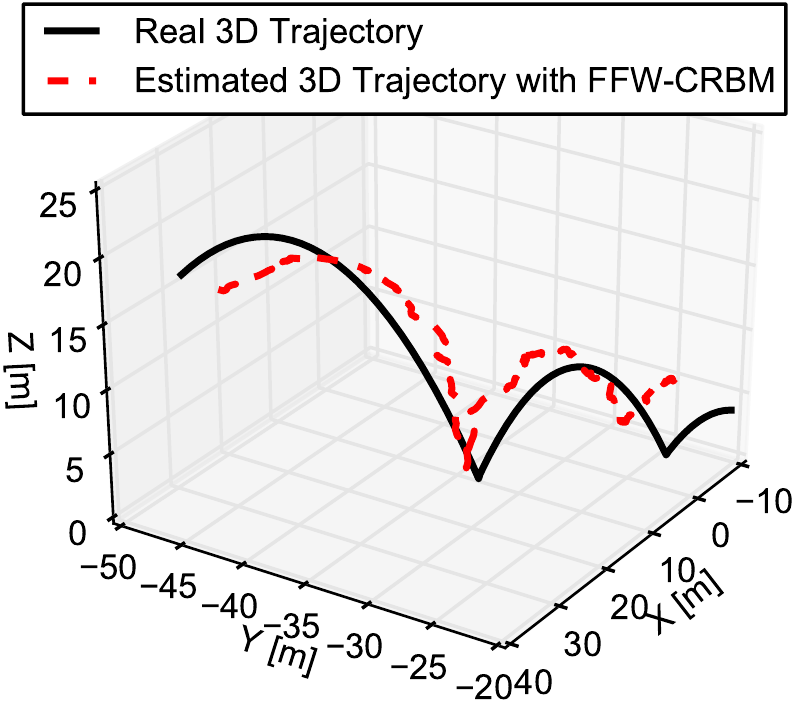}
   \includegraphics[width=0.19\linewidth]{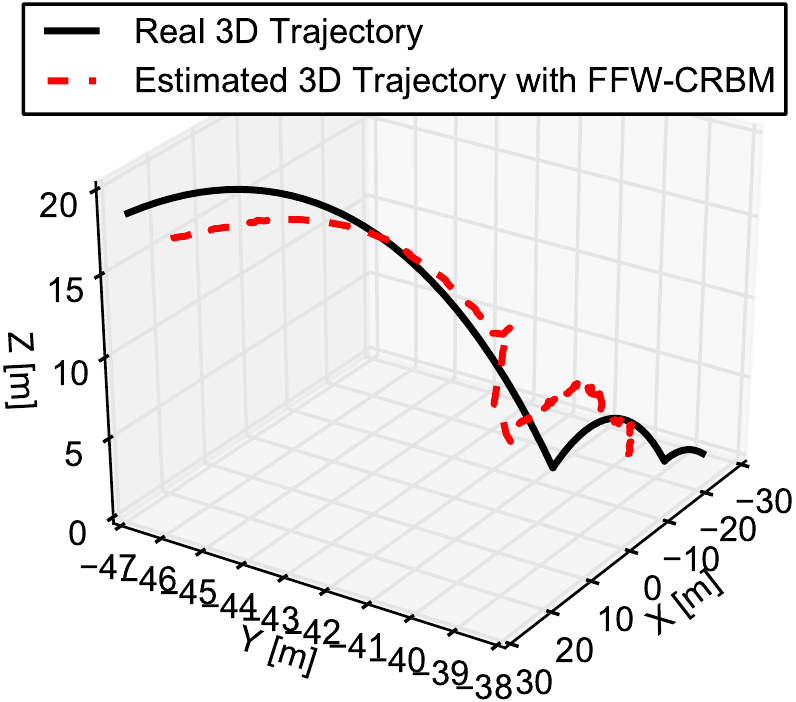}
   \includegraphics[width=0.19\linewidth]{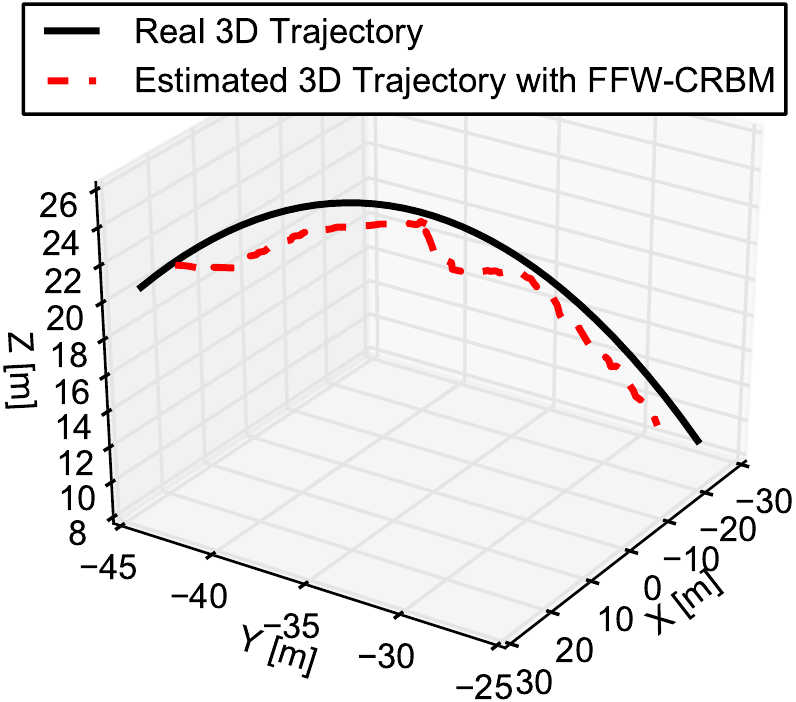}   
\end{center}
\caption{Estimation of different 3D balls trajectories from their 2D counterparts with DFFW-CRBM (top) and FFW-CRBM (bottom) showing that our method outperforms state-of-the-art techniques while requiring less data.}
\label{fig:diffballstrajectories}
\end{figure*}

\begin{figure}[t]
\begin{center}
\begin{minipage}[l]{0.49\columnwidth}
   \includegraphics[width=0.9\linewidth]{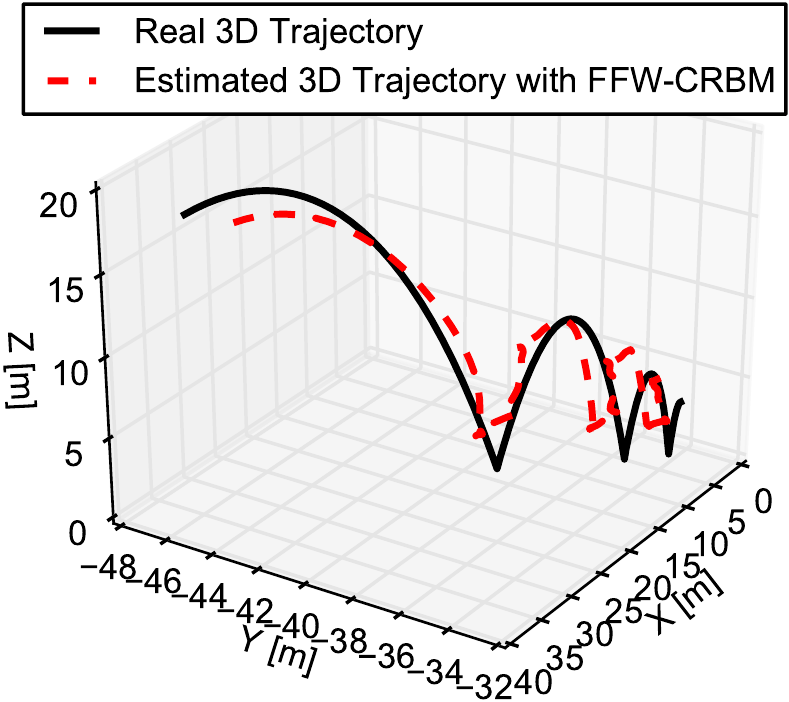}
\end{minipage}
\begin{minipage}[r]{0.49\columnwidth}
   \includegraphics[width=0.93\linewidth]{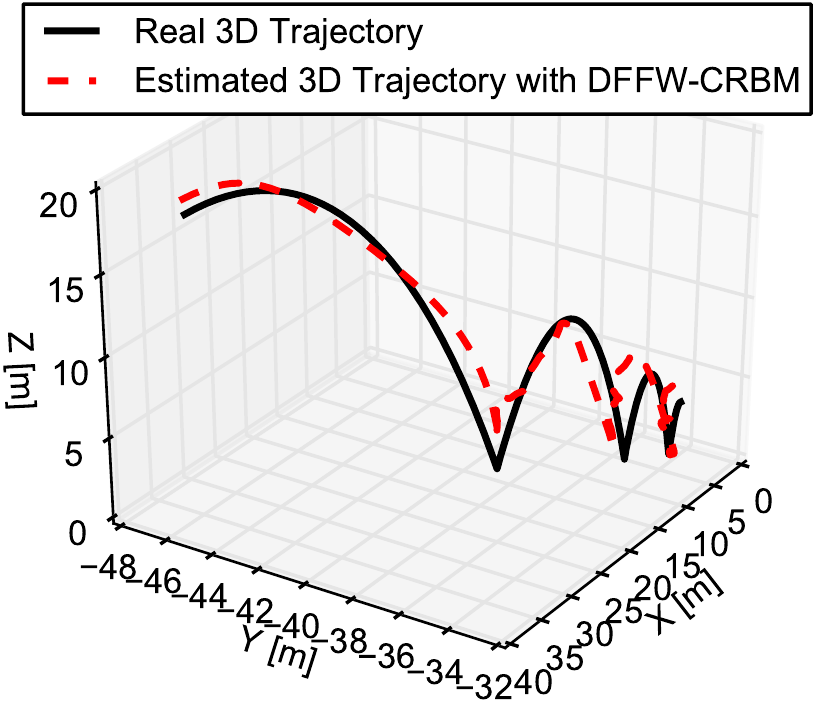}
\vspace{0.37em}
\end{minipage}
\begin{minipage}[l]{0.49\columnwidth}
   \includegraphics[width=0.9\linewidth]{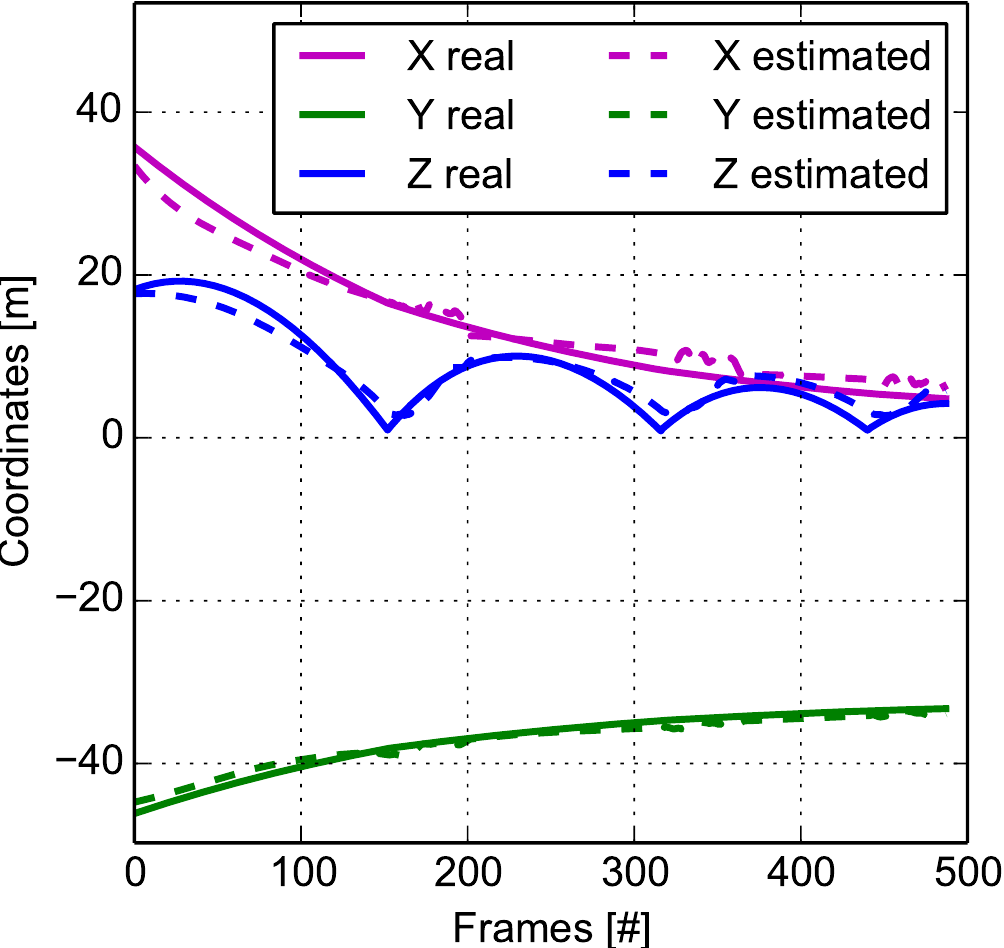}
\end{minipage}
\begin{minipage}[r]{0.49\columnwidth}
   \includegraphics[width=0.9\linewidth]{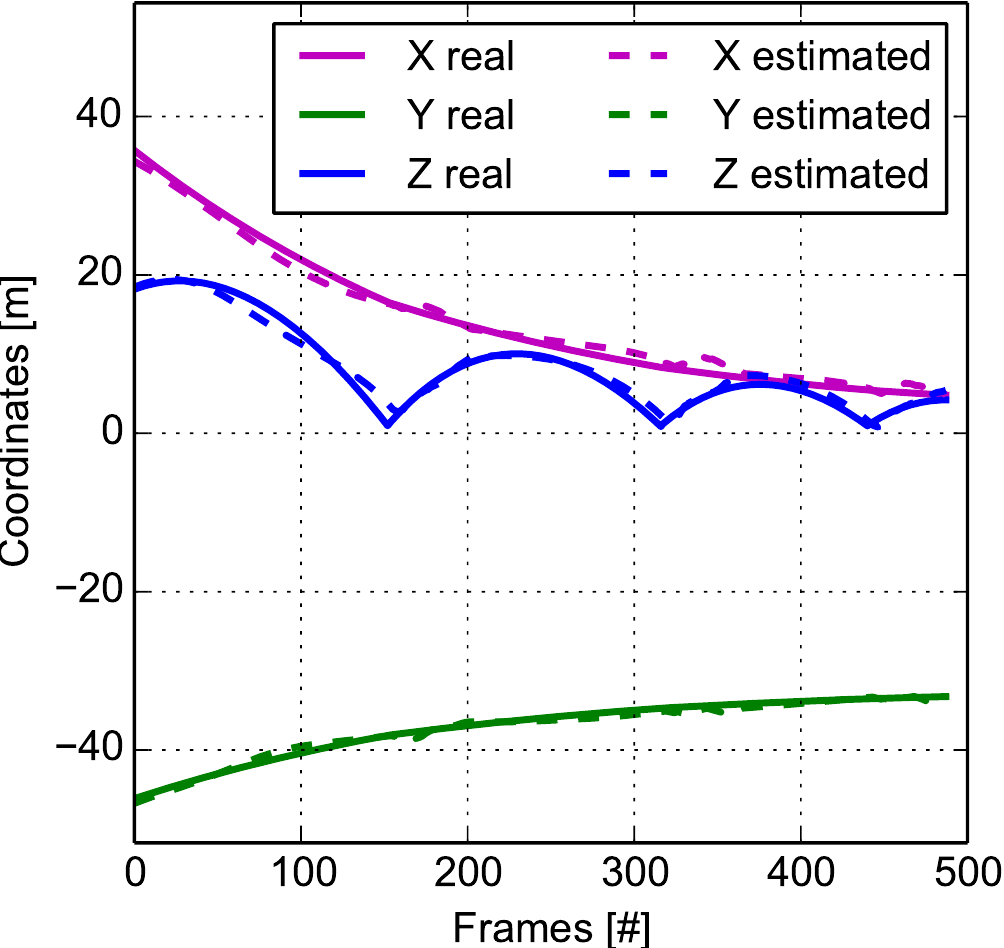}
\end{minipage}
\end{center}
\caption{Estimation of the 3D trajectory for the center of one ball from its 2D projection using FFW-CRBM (left) and DFFW-CRBM (right). The top figure presents the trajectory in the 3D space, while the bottom figure presents the Ox, Oy, Oz coordinates of the same trajectory in a 2D plot.}
\label{fig:ballffwcrbm}
\end{figure}

\begin{figure}[t]
\begin{center}
   \includegraphics[width=0.6\linewidth]{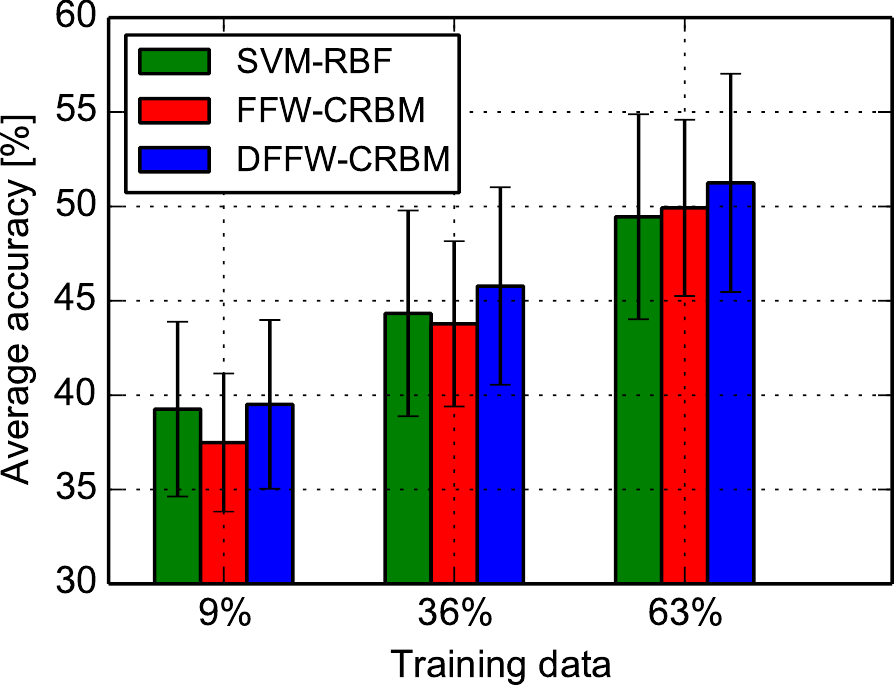}
\end{center}
\caption{Average classification accuracies with mean and standard deviation, over all balls trajectories, when the amount of training data is increased.}
\label{fig:increaseaccballs}
\end{figure}

Figures~\ref{fig:diffballstrajectories} and~\ref{fig:ballffwcrbm} compare the capabilities of DFFW-CRBMs on estimating different 3D trajectories of balls picked at random to FFW-CRBMs, showing that our method is capable of achieving closely correlated transitions to the real trajectory. Interestingly, DFFW-CRBMs can handle discontinuities ``less abruptly'' compared to FFW-CRBMs. The cross-validation results showing the performance of all models of all ball trajectories are summarized in Table~\ref{tab:claspreddiffballs}. In terms of classification, SVM-RBF, FFW-CRBM, and DFFW-CRBM perform almost similarly, with a slightly advantage of DFFW-CRBM\footnote{It is worth noting that in this scenario the random guess for classification would have an accuracy of $25\%$.}. In the case of 3D coordinates estimation from 2D projection at a time-step $t$, DFFW-CRBM clearly outperforms state-of-the-art methods with a NRMSE almost twice smaller than FCRBMs and FFW-CRBMs. Besides that, in this case, the mean value of the correlation coefficient for DFFW-CRBM is $0.62$, double than that for FFW-CRBM, while the one for FCRBM is powerless (i.e below zero). For the multi-step prediction of near-future 3D point coordinates, DFFW-CRBM has an even more significant improvement. It is worth highlighting that in this scenario, the average PCC value after one step prediction is almost perfectly $0.95$, while after 50 steps predicted into the future the mean PCC value is still positive and larger than those of the other methods. In a final set of experiments we tested the change in the accuracy of classification as a number of data points used. These are summarized in the bar-graph in Figure~\ref{fig:increaseaccballs}, showing that our method slightly outperforms the state-of-the-art techniques in all cases. 
\subsection{Human Activity  Recognition}
\label{subsec:har}
Given the above successes, next we evaluate the performance of our method on real-world data representing a variety of human activities.  In each set of experiments, we targeted two main objectives and a third secondary one. The first two corresponded to estimating three-dimensional joint coordinates from two-dimensional projections as well as predicting such coordinates in near future, while the third involved classifying activities based on only two-dimensional joint coordinates. Please note that the third experiment is exceptionally hard due to the loss of three-dimensional information making different activities more similar. 

\textbf{Human 2.6m dataset.} For all experiments, we used the real-world comprehensive benchmark database~\cite{IonescuSminchisescu11,h36m_pami}, containing 17 activities performed by 11 professional actors (6 males and 5 females) with over 3.6 million 3D human poses and their corresponding images. Further, for 7 actors, the database accurately reports 32 human skeleton joint positions in 3D space, together with their 2D projections acquired at 50 frames per seconds (FPS). 

\begin{figure}[t]
\begin{center}
   \includegraphics[width=0.6\linewidth]{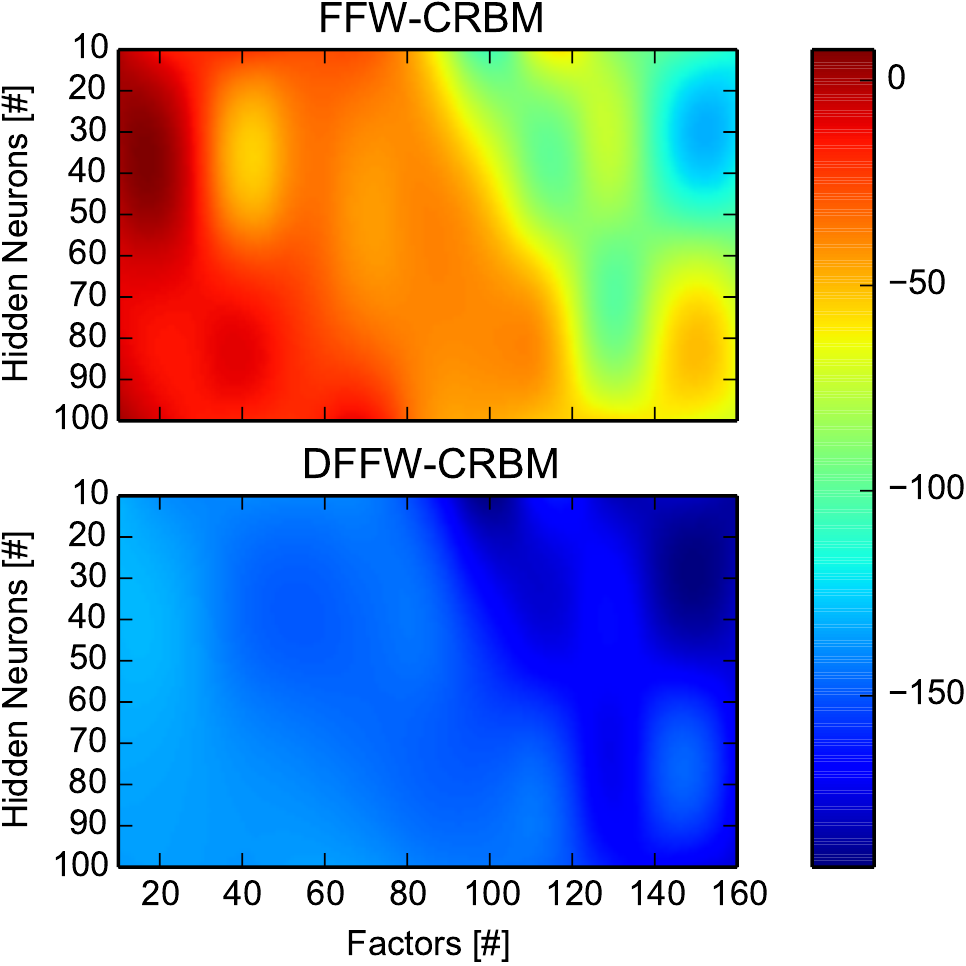}
\end{center}
\caption{Averaged energy levels of FFW-CRBM and DFFW-CRBM for the human activities experiments when the parameters (i.e. number of hidden neurons and factors) are varying. The training was done for 100 epochs.}
\label{fig:enleveltuneparameter}
\end{figure}

We used these seven actors being Subject 1 (S1), Subject 5 (S5), Subject 6 (S6), Subject 7 (S7), Subject 8 (S8), Subject 9 (S9), Subject 11 (S11) accompanied with their corresponding joint activities, such as Purchasing (A1), Smoking (A2), Phoning (A3), Sitting-Down (A4), Eating (A5), Walking-Together (A6), Greeting (A7), Sitting (A8), Posing (A9), Discussing (A10), Directing (A11), Walking (A12), and  Waiting (A13). To avoid computational overhead, we have also reduced the temporal resolution of the data to 5 FPS leading to a total of 46446 training and testing instances. The instances were split between different subjects as: S1 (5514 instances), S5 (8748 instances), S6 (5402 instances),S7 (9081 instances), S8 (5657 instances), S9 (6975 instances), and S11 (5069 instances).

\textbf{Deep Learner Setting:} The visible layers of both the FFW-CRBM and DFFW-CRBM were set to 160 neurons corresponding to 96 neurons for the 3D coordinates of the joints, and 64  for their 2D projections at time $t$. The label layer consisted of 13 neurons (one for each of the activities), and the history layers included 320 neurons corresponding to 5 history frames each incorporating 2D joint coordinates. The size of the hidden layer was set to $10$ neurons, and the number of factors to $100$, as explained in the next paragraph. Furthermore, a learning rate of $10^{-5}$ was used to guarantee bounded reconstruction errors. The number of the Markov Chain steps in the training phase and of the Gibbs sampling in the testing phase were set to 3, and the weights were initialized with $\mathcal{N}(0,0.3)$. Further particularities, such as momentum and weight decay were set to $0.5$ and $0.0002$. Also, all data were normalized to have a 0 mean and unit standard deviation. 

\textbf{Importance of disjunctive Factoring:} Similarly with the previous experiment on simulated balls trajectories, we searched for the optimal number of hidden neurons and factors, by performing exhaustive search and varying the number of hidden neurons and factors from 10 to 100 and from 10 to 160, respectively. Figure~\ref{fig:enleveltuneparameter} depicts  on the same scale the averaged energy levels for both FFW-CRBM and DFFW-CRBM, after being trained for 100 epochs. As before, in the balls experiment, the energy levels of both models are more affected by the number of factors than the number of hidden neurons. Even if we are scrutinizing unnormalized energy levels, the fact that the energy levels of DFFW-CRBM are always much lower than the energy levels of FFW-CRBM reflects the importance of the disjunctive factoring. By quantifying and averaging all the energy levels for each model, we may observe that DFFW-CRBM has in average approximately three times less energy than FFW-CRBM (i.e. $-153.32\pm17.31$ for DFFW-CRBM, and $-48.57\pm33.92$ for FFW-CRBM).

\subsubsection{Training and Testing on The Same Person}
Here, data from the same subject has been used for both training and testing. Emulating real-world 3D trajectory prediction settings where labeled data is scarce, we made use of only 10$\%$ of the available data for training and 90$\%$ for testing with the aim of performing accurate one and multi-step 3D trajectory predictions. 

Results in Tables~\ref{tab:accsameperson},~\ref{tab:onepredsameperson}, and~\ref{tab:multipredsameperson} show that DFFW-CRBMs are capable of achieving better performance than state-of-the-art techniques in both classification and prediction even when only using a small amount of training data. These results provide a proof-of-concept to the fact that DDFW-CRBMs are capable of accurately predicting (in both one-step and multi-step scenarios) 3D trajectories from their 2D projections by using only 10$\%$ of the data for training and 90$\%$ for testing.
\begin{table}
\scriptsize
\begin{center}
\begin{tabular}{|c|c|c|c|}
\hline
Persons&SVM-RBF&FFW-CRBM&DFFW-CRBM \\
\hline\hline
S1 & 49.77 &50.53&49.34\\
S5 & 36.92 &38.82&40.21\\
S6 & 30.68&31.51&30.18\\
S7 & 38.50&37.03&37.94\\
S8 & 26.49&30.41&31.32\\
S9 & 24.69&28.12&22.63\\
S11 & 34.56&34.21&32.16\\
\hline
Average &34.51&\textbf{35.80}&34.83\\
\hline
\end{tabular}
\end{center}
\caption{Classification accuracies in percentages for the human activities experiments, when training and testing data belong to the same person.}
\label{tab:accsameperson}
\end{table}

\begin{table*}
\tiny
\tabcolsep=0.07cm
\begin{center}
\begin{tabular}{|c|r|r|r|r|r|r|r|r|r|}
\hline
\multirow{2}{*}{Persons}&\multicolumn{3}{c|}{FCRBM}&\multicolumn{3}{c|}{FFW-CRBM}&\multicolumn{3}{c|}{DFFW-CRBM} \\
\cline{2-10}
&NRMSE [\%]&PCC&P-value&NRMSE [\%]&PCC&P-value&NRMSE [\%]&PCC&P-value\\
\hline\hline
S1 &8.41$\pm$3.75&0.02$\pm$0.10&0.49$\pm$0.29&9.93$\pm$7.47&0.13$\pm$0.37&0.15$\pm$0.26&6.36$\pm$3.45&0.54$\pm$0.29&0.05$\pm$0.16\\
S5 &6.70$\pm$2.44&-0.03$\pm$0.09&0.54$\pm$0.28&6.95$\pm$3.21&0.10$\pm$0.33&0.16$\pm$0.26&4.30$\pm$2.30&0.68$\pm$0.25&0.02$\pm$0.10\\
S6 &4.41$\pm$1.93&0.03$\pm$0.09&0.53$\pm$0.28&4.50$\pm$2.37&0.01$\pm$0.28&0.21$\pm$0.29&3.19$\pm$1.64&0.50$\pm$0.32&0.05$\pm$0.16\\
S7 &9.14$\pm$3.46&0.02$\pm$0.10&0.49$\pm$0.29&9.16$\pm$4.30&0.13$\pm$0.35&0.14$\pm$0.25&6.19$\pm$3.11&0.71$\pm$0.24&0.01$\pm$0.09\\
S8 &8.31$\pm$3.37&-0.00$\pm$0.11&0.47$\pm$0.29&8.23$\pm$4.42&0.02$\pm$0.27&0.26$\pm$0.31&4.96$\pm$2.57&0.62$\pm$0.26&0.05$\pm$0.18\\
S9 &7.25$\pm$2.74&0.00$\pm$0.09&0.55$\pm$0.28&8.40$\pm$5.05&0.01$\pm$0.27&0.22$\pm$0.29&4.63$\pm$2.34&0.54$\pm$0.32&0.05$\pm$0.17\\
S11 &9.62$\pm$4.05&-0.00$\pm$0.10&0.54$\pm$0.28&9.89$\pm$5.94&0.06$\pm$0.32&0.15$\pm$0.25&6.82$\pm$3.82&0.53$\pm$0.35&0.04$\pm$0.14\\
\hline
Average &$\approx$7.69$\pm$3.10&$\approx$0.01$\pm$0.09&$\approx$0.51$\pm$0.28&$\approx$8.15$\pm$4.68&$\approx$0.07$\pm$0.31&$\approx$0.18$\pm$0.27&\textbf{$\approx$5.21$\pm$2.75}&\textbf{$\approx$0.59$\pm$0.29}&\textbf{$\approx$0.04$\pm$0.14}\\
\hline
\end{tabular}
\end{center}
\caption{The 3D estimation of the human joints from their 2D counterpart at the present time, when the training and testing are done on the same person. The results are presented with mean and standard deviation.}
\label{tab:onepredsameperson}
\end{table*}

\begin{table*}
\tiny
\tabcolsep=0.07cm
\begin{center}
\begin{tabular}{|c|c|r|r|r|r|r|r|r|r|r|}
\hline
Steps&\multirow{2}{*}{Persons}&\multicolumn{3}{c|}{FCRBM}&\multicolumn{3}{c|}{FFW-CRBM}&\multicolumn{3}{c|}{DFFW-CRBM}\\
\cline{3-11}
Predicted&&NRMSE [\%]&PCC&P-value&NRMSE [\%]&PCC&P-value&NRMSE [\%]&PCC&P-value\\
\hline\hline
&S1 &7.68$\pm$3.71&0.02$\pm$0.09&0.51$\pm$0.28&7.78$\pm$4.95&0.17$\pm$0.32&0.17$\pm$0.27&5.91$\pm$3.45&0.55$\pm$0.26&0.06$\pm$0.16\\
&S5 &6.72$\pm$2.48&-0.05$\pm$0.09&0.50$\pm$0.28&6.41$\pm$2.70&0.19$\pm$0.37&0.18$\pm$0.29&3.88$\pm$1.68&0.68$\pm$0.25&0.01$\pm$0.08\\
After&S6 &4.40$\pm$2.17&0.04$\pm$0.09&0.52$\pm$0.28&4.27$\pm$2.31&0.11$\pm$0.31&0.21$\pm$0.30&3.17$\pm$1.83&0.48$\pm$0.32&0.05$\pm$0.16\\
1 step&S7 &9.07$\pm$3.20&0.02$\pm$0.11&0.46$\pm$0.29&8.78$\pm$3.52&0.27$\pm$0.35&0.06$\pm$0.17&6.52$\pm$3.05&0.73$\pm$0.17&0.00$\pm$0.02\\
&S8 &7.16$\pm$3.08&0.01$\pm$0.12&0.48$\pm$0.31& 6.42$\pm$3.51&0.04$\pm$0.23&0.30$\pm$0.31&3.93$\pm$1.87&0.69$\pm$0.19&0.01$\pm$0.05\\
&S9 &6.98$\pm$2.64&-0.01$\pm$0.09&0.54$\pm$0.29&6.91$\pm$3.17&0.08$\pm$0.26&0.22$\pm$0.28&4.40$\pm$1.69&0.64$\pm$0.20&0.01$\pm$0.08\\
&S11 &9.55$\pm$4.05&-0.00$\pm$0.08&0.56$\pm$0.25&8.92$\pm$4.66&0.10$\pm$0.31&0.18$\pm$0.27&7.02$\pm$4.00&0.51$\pm$0.44&0.04$\pm$0.14\\
\cline{2-11}
&Average&$\approx$7.37$\pm$3.05&$\approx$0.00$\pm$0.09&$\approx$0.51$\pm$0.28&$\approx$7.07$\pm$3.55&$\approx$0.14$\pm$0.31&$\approx$0.18$\pm$0.27&\textbf{$\approx$4.96$\pm$2.51}&\textbf{$\approx$0.61$\pm$0.26}&\textbf{$\approx$0.03$\pm$0.1}\\
\hline\hline
&S1&10.88$\pm$3.17&0.01$\pm$0.10&0.55$\pm$0.32&10.23$\pm$5.39&0.10$\pm$0.41&0.17$\pm$0.26&8.22$\pm$4.56&-0.03$\pm$0.29&0.14$\pm$0.24\\
&S5&7.50$\pm$2.20&0.02$\pm$0.10&0.53$\pm$0.28&8.40$\pm$3.20&0.01$\pm$0.43&0.19$\pm$0.30&6.86$\pm$2.47&0.12$\pm$0.43&0.08$\pm$0.19\\
After&S6&5.38$\pm$1.92&0.01$\pm$0.12&0.45$\pm$0.29&4.77$\pm$2.65&-0.02$\pm$0.24&0.22$\pm$0.30&4.44$\pm$2.17&0.12$\pm$0.32&0.12$\pm$0.23\\
50 steps&S7&11.07$\pm$3.61&-0.03$\pm$0.09&0.52$\pm$0.30&10.68$\pm$4.04&0.11$\pm$0.44&0.15$\pm$0.26&9.31$\pm$3.60&0.17$\pm$0.33&0.12$\pm$0.25\\
&S8&15.41$\pm$1.66&0.01$\pm$0.11&0.45$\pm$0.29&11.33$\pm$5.81&0.09$\pm$0.24&0.26$\pm$0.29&9.91$\pm$5.73&0.08$\pm$0.38&0.10$\pm$0.21\\
&S9&9.25$\pm$2.10&0.01$\pm$0.11&0.48$\pm$0.28&8.56$\pm$3.23&0.03$\pm$0.25&0.25$\pm$0.28&7.48$\pm$3.60&-0.02$\pm$0.39&0.12$\pm$0.22\\
&S11&14.39$\pm$2.71&-0.01$\pm$0.09&0.52$\pm$0.28&10.93$\pm$5.78&0.17$\pm$0.37&0.18$\pm$0.29&8.56$\pm$4.47&0.05$\pm$0.51&0.12$\pm$0.26\\
\cline{2-11}
&Average&$\approx$10.55$\pm$2.48&$\approx$0.00$\pm$0.10&$\approx$0.5$\pm$0.29&$\approx$9.27$\pm$4.3&$\approx$0.07$\pm$0.34&$\approx$0.20$\pm$0.28&\textbf{$\approx$7.82$\pm$3.8}&$\approx$0.07$\pm$0.37&\textbf{$\approx$0.11$\pm$0.23}\\
\hline
\end{tabular}
\end{center}
\caption{Multi-step 3D prediction for the human activities experiments, when the training and testing are done on the same person. The results are presented with mean and standard deviation.}
\label{tab:multipredsameperson}
\end{table*}

\begin{figure}[t]
\begin{center}
   \includegraphics[width=0.6\linewidth]{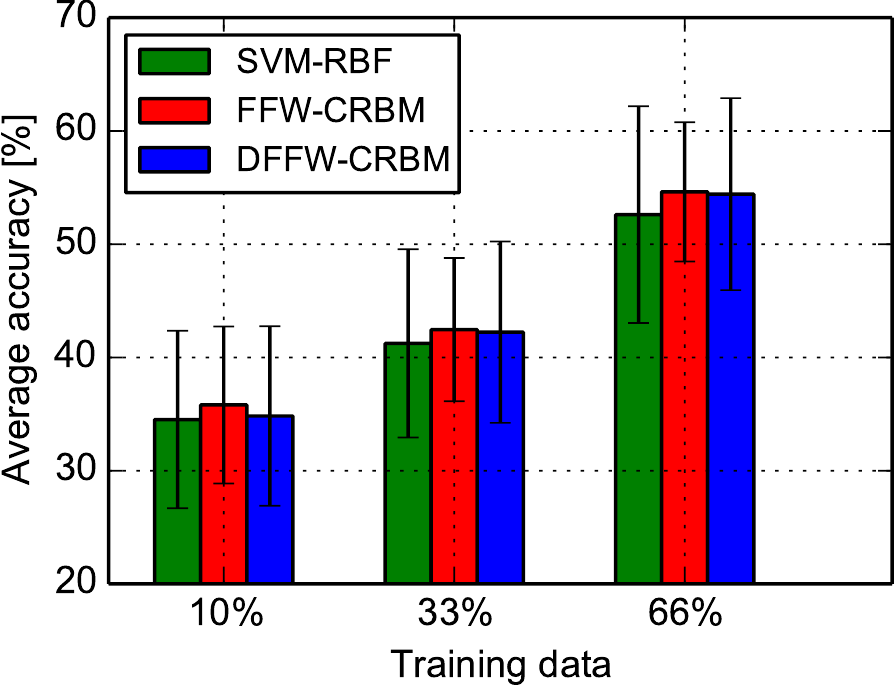}
\end{center}
\caption{Average classification accuracies with mean and standard deviation for the human activities experiments, over all subjects, when the data for training and testing the models come from the same person and the amount of training data is increased.}
\label{fig:increaseacc}
\end{figure}

\textbf{Activity Recognition (Classification)}
The goal in this set of experiments was to classify the 13 activities based on only their 2D projections. Please note that such a task is substantially difficult to solve due to the loss of information exhibited by the performed projection. Namely, activities different in 3D space might resemble high similarities in their 2D projections leading to low classification accuracies. Table~\ref{tab:accsameperson} reports the accuracy performance of DFFW-CRBMs, against state-of-the-art methods including SVMs and FFW-CRBMs. By averaging the results over all subjects, we can observe that all three models perform comparable. It is worth mentioning that the classification accuracy for random choice in this scenario would be $7.69\%$ and all models performs approximately 5 times better. 

\begin{figure}[t]
\begin{center}
   \includegraphics[width=0.6\linewidth]{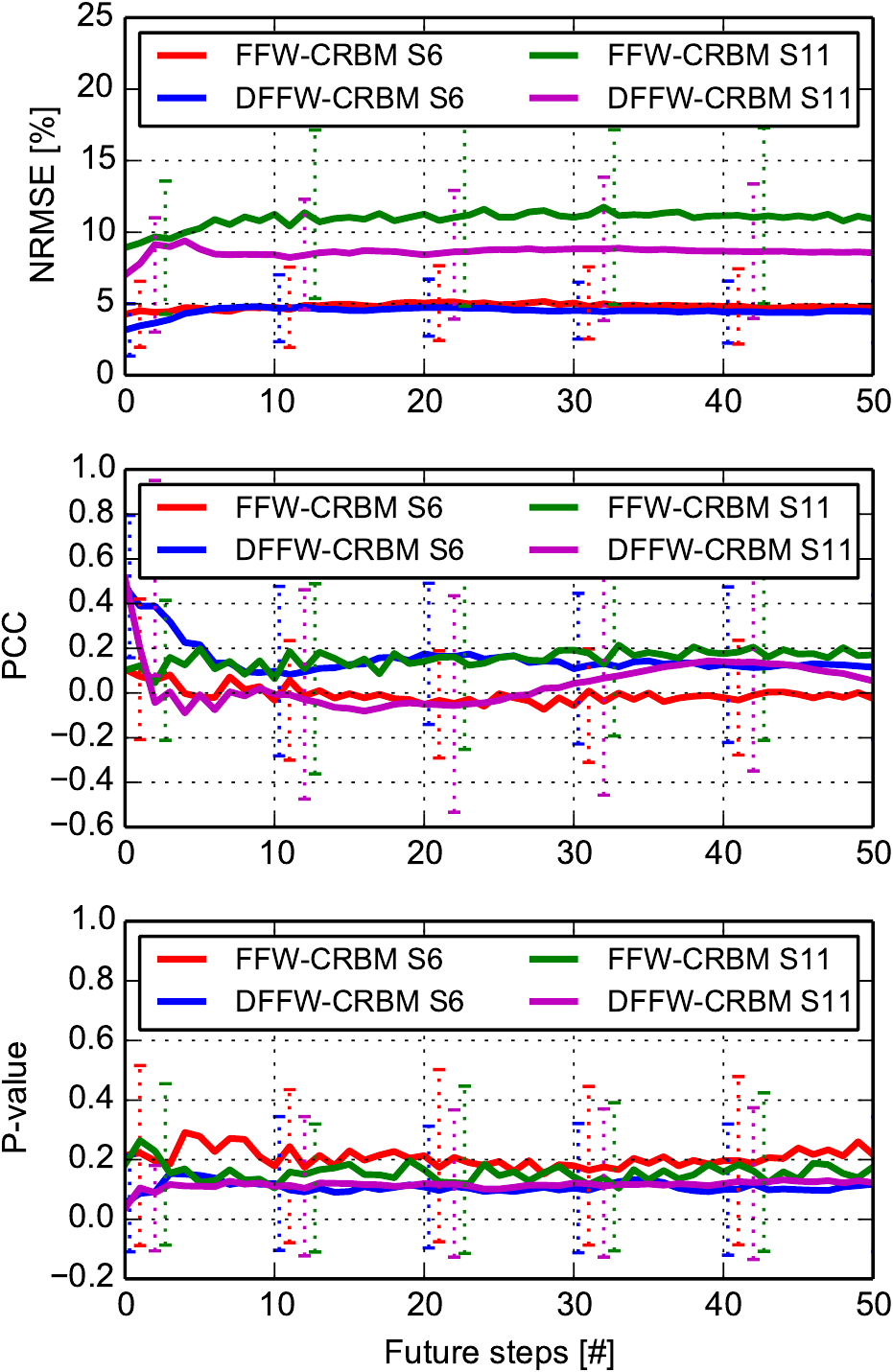}
\end{center}
\caption{Multi-step 3D prediction on the worst performer (S11) and best performer (S6) subjects, when the data for training and testing the models come from the same person.}
\label{fig:m3ds8}
\end{figure}
We also performed two more experiments to classify activities with more input data points to prove the correctness of the presented methods and show DFFW-CRBMs is capable of achieving state-of-the-art classification results. Here, we used 33$\%$ and 66$\%$ of the data to train the models, and the remaining for test. It is clear from Figure~\ref{fig:increaseacc} that all models increase in performance as the amount of training data increases, reaching around 55$\%$ accuracy when 66$\%$ of the data is used for training.
\begin{figure}[t]
\begin{center}
   \includegraphics[width=0.6\linewidth]{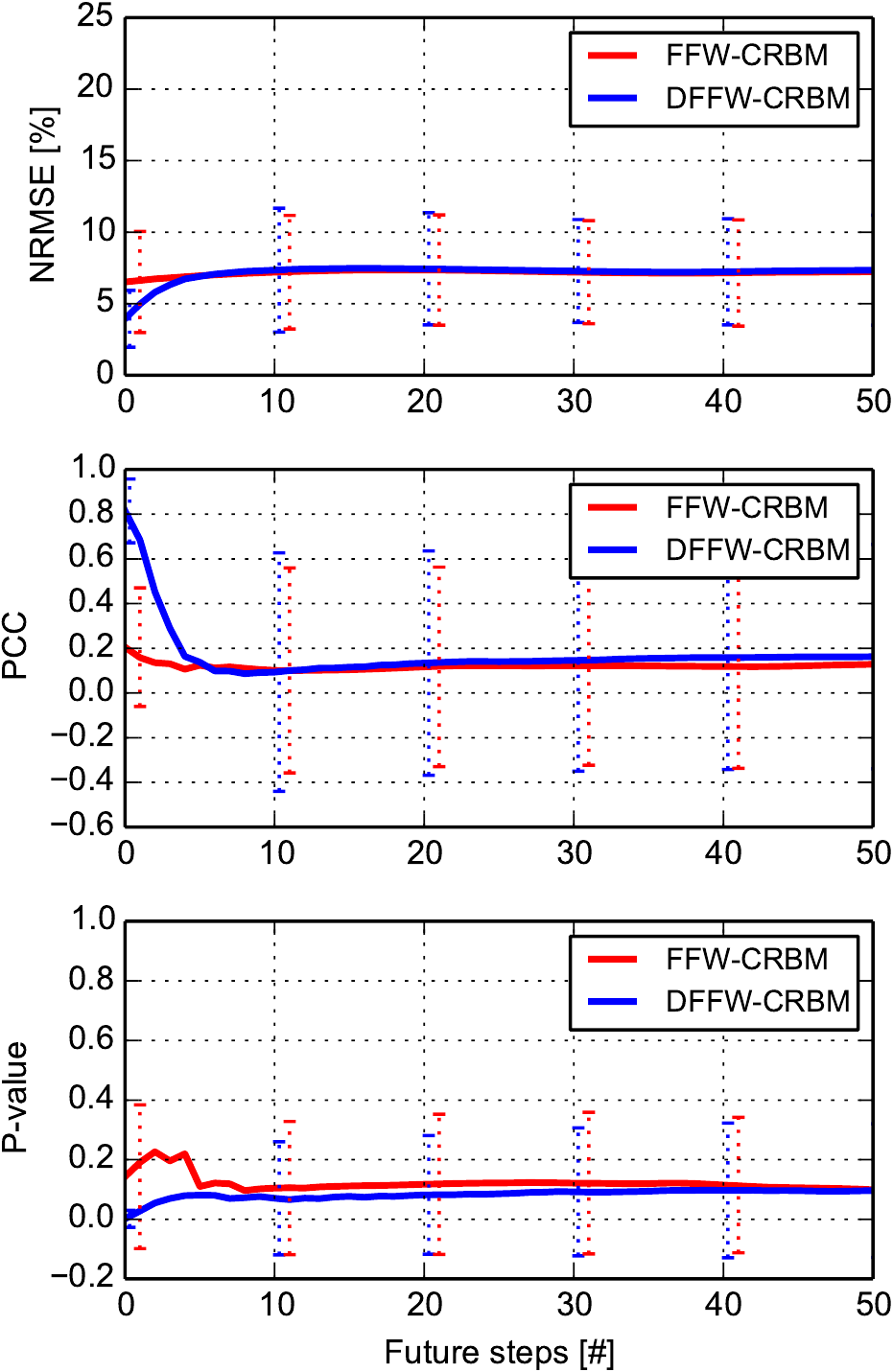}
\end{center}
\caption{Multi-step 3D prediction using cross-validation on all subjects, when the data for training and testing the models come from different persons.}
\label{fig:m3cv}
\end{figure}

\textbf{Estimating 3D Skeleton Coordinates from 2D Projections (Present Step Prediction)}
In this task, we estimate the 3D joint coordinates from their 2D counterpart while using 10$\%$ training data. Results depicted in Table~\ref{tab:onepredsameperson} show that DFFW-CRBMs achieves better performance than FFW-CRBMs and FCRBM. Though FFW-CRBMs perform comparatively, it is worth noting that the PCC and P-values signify the fact that DFFW-CRBMs drastically outperform FFW-CRBMs in the sense that the predictions are correlated with ground truth, a property essential for accurate and reliable predictions.   

\textbf{Prediction of 3D Skeleton Trajectories (Multi-Step Prediction)}
Here, the goal was to perform multi-step predictions of the 3D skeleton joints based on only 2D projections. Starting from a 2D initial state, the model was executed autonomously by recursively feeding-back 2D outputs to perform next-step predictions. Definitely, the performance is expected to degrade since the prediction errors accumulate with time. 
Table~\ref{tab:multipredsameperson}, showing the performance of the models after 1 and 50 step predictions, validate this phenomenon since all metrics show a decrease in both models' performance over time. Table~\ref{tab:multipredsameperson}, however, also signify that DFFW-CRBMs outperform FFW-CRBMs in both one and multi-step predictions achieving an average NRMSE of 7.82 compared to 9.27 NRMSE for FFW-CRBMs. Further results are summarized by Figure~\ref{fig:m3ds8} showing the minimum and maximum performance results of both models. In these experiments, clearly, DFFW-CRBM is the best performer in both, prediction errors and correlations.   

\subsubsection{Testing Generalization Capabilities}
\textbf{Motivation:} In the second set of human activities experiments, our goal was to determine to what extend can DFFW-CRBMs generalize across different human subjects and activities. The main motivation is that in reality subject-specific data is scarce, while data available from different users or domains is abundant.  Results reported in Table~\ref{tab:claspreddiffperson} and Figure~\ref{fig:m3cv} show that DFFW-CRBMs are capable of generalizing beyond specific subjects due to their ability in learning latent features shared among a variety of tasks. 
\begin{table}
\scriptsize
\tabcolsep=0.03cm
\begin{center}
\begin{tabular}{|c|c|c|c|r|r|r|}
\hline
\multicolumn{2}{|c|}{\multirow{2}{*}{Task}}&\multirow{2}{*}{Metrics}&\multicolumn{4}{c|}{Methods}\\
\cline{4-7}
\multicolumn{2}{|c|}{}&&SVM-RBF&FCRBM&FFW-CRBM&DFFW-CRBM \\
\hline\hline
\multicolumn{2}{|c|}{Classification}&Accuracy[\%]&37.93$\pm$5.04&N/A&\textbf{44.96$\pm$2.68}&44.49$\pm$6.60\\
\hline\hline
\multicolumn{2}{|c|}{Present Step}&NRMSE[\%]&N/A&7.58$\pm$3.62&7.52$\pm$3.63&\textbf{3.93$\pm$1.75}\\
\multicolumn{2}{|c|}{3D estimation}&PCC&N/A&-0.00$\pm$0.09&0.14$\pm$0.24&\textbf{0.79$\pm$0.16}\\
\multicolumn{2}{|c|}{}&P-value&N/A&0.52$\pm$0.28&0.21$\pm$0.28&\textbf{0.01$\pm$0.03}\\
\hline\hline
&After&NRMSE[\%]&N/A&6.60$\pm$3.53&6.52$\pm$3.54&\textbf{3.95$\pm$1.99}\\
&1 step&PCC&N/A&-0.01$\pm$0.11&0.21$\pm$0.27&\textbf{0.81$\pm$0.14}\\
Multi-Step&&P-value&N/A&0.49$\pm$0.29&0.14$\pm$0.24&\textbf{0.01$\pm$0.03}\\
\cline{2-7}
3D&After&NRMSE[\%]&N/A&7.27$\pm$3.81&\textbf{7.24$\pm$3.84}&7.34$\pm$3.84\\
prediction&50 steps&PCC&N/A&0.01$\pm$0.11&0.13$\pm$0.46&\textbf{0.16$\pm$0.50}\\
&&P-value&N/A&0.49$\pm$0.31&0.10$\pm$0.20&0.10$\pm$0.22\\
\hline
\end{tabular}
\end{center}
\caption{Classification, present step 3D estimation, and multi-step 3D prediction, for the human activities experiments, when the training and the testing are done on different persons. The results are cross-validated and presented with mean and standard deviation.}
\label{tab:claspreddiffperson}
\end{table}

\textbf{Experiments:} Here, data from 6 subjects was used to train the models, and predictions on an unseen subject were performed. The procedure was then repeated to cross-validate the results. Further, to emulate real-world settings only 10$\%$ of the data was used for training. During testing, however, all data from the all testing subjects was used increasing the tasks' difficulty. The same three goals of the previous experiments were targeted.

\textbf{Activity Recognition (Classification):} 
Results reported in Table~\ref{tab:claspreddiffperson} show that DFFW-CRBMs achieve comparable results to FFW-CRBMs at an accuracy of 44.5 $\%$ both outperforming SVMs. Clearly, these classification results resemble higher accuracies when compared to these in Table~\ref{tab:accsameperson}. The reasons can be attributed back to the availability of similar domain data from other subjects signifying the latent feature similarities automatically learn by DFFW-CRBMs. 

\textbf{Estimation of 3D Skeleton Coordinates from 2D Projections (Present Step Prediction):} Again, DFFW-CRBMs achieve better performance  than FFW-CRBMs in present step estimation of the 3D skeleton joints from 2D projections, while both outperform FCRBM. It is worth highlighting that DFFW-CRBMs are capable of attaining a high average prediction correlation to ground-truth of almost 0.8.

\textbf{Prediction of 3D skeleton Trajectories (Multi-Step Prediction):} Finally, Figure~\ref{fig:m3cv} shows that DFFW-CRBMs are capable of surpassing FFW-CRBMs in multi-step predictions on unseen subjects achieving low prediction errors and high ground truth correlation.  

\section{Conclusion}
\label{Sec:Conclusions}
In this paper we proposed \textit{disjunctive factored four-way conditional restricted Boltzmann machines} (DFFW-CRBMs). These novel machine learning techniques can be used for estimating 3D trajectories from their 2D projections using limited amounts of labeled data. 
Due to the new tensor factoring introduced by DFFW-CRBMs, these machines are capable of achieving substantially lower energy levels than state-of-the-art techniques leading to more accurate predictions and classification results. Furthermore, DFFW-CRBMs are capable of performing classification and accurate near-future predictions simultaneously in one unified framework.

Two sets of experiments, one on a simulated ball trajectories dataset and one on a real-world benchmark database, demonstrate the effectiveness of DFFW-CRBMs. The empirical evaluation showed that our methods are capable of outperforming state-of-the-art machine learning algorithms in both classification and regression. Precisely, DFFW-CRBM were capable of achieving substantially lower energy levels (approximately three times less energy on the overall datasets, independently on the number of factors or hidden neurons) than FFW-CRBM. This leads to at least double accuracies for real-valued predictions, while acquiring similar classification performance, at no increased computational complexity costs.


\bibliography{refs}

\begin{thebibliography}{10}
\expandafter\ifx\csname url\endcsname\relax
  \def\url#1{\texttt{#1}}\fi
\expandafter\ifx\csname urlprefix\endcsname\relax\def\urlprefix{URL }\fi
\expandafter\ifx\csname href\endcsname\relax
  \def\href#1#2{#2} \def\path#1{#1}\fi

\bibitem{3dtrajrobotics}
H.~Silva, A.~Dias, J.~Almeida, A.~Martins, E.~Silva, Real-time 3d ball
  trajectory estimation for robocup middle size league using a single camera,
  in: T.~Röfer, N.~Mayer, J.~Savage, U.~Saranlı (Eds.), RoboCup 2011: Robot
  Soccer World Cup XV, Vol. 7416 of Lecture Notes in Computer Science, Springer
  Berlin Heidelberg, 2012, pp. 586--597.
\newblock \href {http://dx.doi.org/10.1007/978-3-642-32060-6_50}
  {\path{doi:10.1007/978-3-642-32060-6_50}}.

\bibitem{3dtrajmedicie}
S.~Qiu, Y.~Yang, J.~Hou, R.~Ji, H.~Hu, Z.~Wang, Ambulatory estimation of 3d
  walking trajectory and knee joint angle using marg sensors, in: Innovative
  Computing Technology (INTECH), 2014 Fourth International Conference on, 2014,
  pp. 191--196.
\newblock \href {http://dx.doi.org/10.1109/INTECH.2014.6927742}
  {\path{doi:10.1109/INTECH.2014.6927742}}.

\bibitem{3dtrajbiology}
S.~Maleschlijski, G.~Sendra, A.~Di~Fino, L.~Leal-Taixé, I.~Thome, A.~Terfort,
  N.~Aldred, M.~Grunze, A.~Clare, B.~Rosenhahn, A.~Rosenhahn, Three dimensional
  tracking of exploratory behavior of barnacle cyprids using stereoscopy,
  Biointerphases 7~(1-4).
\newblock \href {http://dx.doi.org/10.1007/s13758-012-0050-x}
  {\path{doi:10.1007/s13758-012-0050-x}}.

\bibitem{3dtrajdarkmatter}
M.~Jauzac, E.~Jullo, J.-P. Kneib, H.~Ebeling, A.~Leauthaud, C.-J. Ma,
  M.~Limousin, R.~Massey, J.~Richard, A weak lensing mass reconstruction of the
  large-scale filament feeding the massive galaxy cluster macs j0717.5+3745,
  Monthly Notices of the Royal Astronomical Society 426~(4) (2012) 3369--3384.
\newblock \href {http://dx.doi.org/10.1111/j.1365-2966.2012.21966.x}
  {\path{doi:10.1111/j.1365-2966.2012.21966.x}}.

\bibitem{3dtrajwith3cameras}
J.~Tao, B.~Risse, X.~Jiang, R.~Klette, 3d trajectory estimation of simulated
  fruit flies, in: Proceedings of the 27th Conference on Image and Vision
  Computing New Zealand, IVCNZ '12, ACM, New York, NY, USA, 2012, pp. 31--36.
\newblock \href {http://dx.doi.org/10.1145/2425836.2425844}
  {\path{doi:10.1145/2425836.2425844}}.

\bibitem{3dtrajwithradar}
J.~Pinezich, J.~Heller, T.~Lu, Ballistic projectile tracking using cw doppler
  radar, Aerospace and Electronic Systems, IEEE Transactions on 46~(3) (2010)
  1302--1311.
\newblock \href {http://dx.doi.org/10.1109/TAES.2010.5545190}
  {\path{doi:10.1109/TAES.2010.5545190}}.

\bibitem{Larochelle+Bengio-2008}
H.~Larochelle, Y.~Bengio, Classification using discriminative restricted
  boltzmann machines, in: Proceedings of the 25th International Conference on
  Machine Learning, ICML '08, ACM, 2008, pp. 536--543.
\newblock \href {http://dx.doi.org/10.1145/1390156.1390224}
  {\path{doi:10.1145/1390156.1390224}}.

\bibitem{Salakhutdinov07restrictedboltzmann}
R.~Salakhutdinov, A.~Mnih, G.~Hinton, Restricted boltzmann machines for
  collaborative filtering, in: In Machine Learning, Proceedings of the
  Twenty-fourth International Conference (ICML 2004). ACM, AAAI Press, 2007,
  pp. 791--798.

\bibitem{mocanu2014deep}
D.~C. Mocanu, G.~Exarchakos, A.~Liotta, Deep learning for objective quality
  assessment of 3d images, in: Image Processing (ICIP), 2014 IEEE International
  Conference on, 2014, pp. 758--762.
\newblock \href {http://dx.doi.org/10.1109/ICIP.2014.7025152}
  {\path{doi:10.1109/ICIP.2014.7025152}}.

\bibitem{mnih15}
V.~Mnih, K.~Kavukcuoglu, D.~Silver, A.~A. Rusu, J.~Veness, M.~G. Bellemare,
  A.~Graves, M.~Riedmiller, A.~K. Fidjeland, G.~Ostrovski, S.~Petersen,
  C.~Beattie, A.~Sadik, I.~Antonoglou, H.~King, D.~Kumaran, D.~Wierstra,
  S.~Legg, D.~Hassabis, {Human-level control through deep reinforcement
  learning}, Nature 518~(7540) (2015) 529--533.
\newblock \href {http://dx.doi.org/10.1038/nature14236}
  {\path{doi:10.1038/nature14236}}.

\bibitem{ecml2013dec}
H.~Bou-Ammar, D.~C. Mocanu, M.~E. Taylor, K.~Driessens, K.~Tuyls, G.~Weiss,
  Automatically mapped transfer between reinforcement learning tasks via
  three-way restricted boltzmann machines, in: Machine Learning and Knowledge
  Discovery in Databases, Vol. 8189 of Lecture Notes in Computer Science,
  Springer Berlin Heidelberg, 2013, pp. 449--464.
\newblock \href {http://dx.doi.org/10.1007/978-3-642-40991-2_29}
  {\path{doi:10.1007/978-3-642-40991-2_29}}.

\bibitem{Gehler06therate}
P.~V. Gehler, A.~D. Holub, M.~Welling, The rate adapting poisson model for
  information retrieval and object recognition, in: In Proceedings of 23rd
  International Conference on Machine Learning (ICML06), ACM Press, 2006, p.
  2006.

\bibitem{3ddepthnips2014}
D.~Eigen, C.~Puhrsch, R.~Fergus, Depth map prediction from a single image using
  a multi-scale deep network, in: Advances in Neural Information Processing
  Systems 27: Annual Conference on Neural Information Processing Systems 2014,
  December 8-13 2014, Montreal, Quebec, Canada, 2014, pp. 2366--2374.

\bibitem{escalerafacerecognition}
P.~Rasti, T.~Uiboupin, S.~Escalera, G.~Anbarjafari, Convolutional Neural
  Network Super Resolution for Face Recognition in Surveillance Monitoring,
  Springer International Publishing, Cham, 2016, pp. 175--184.
\newblock \href {http://dx.doi.org/10.1007/978-3-319-41778-3_18}
  {\path{doi:10.1007/978-3-319-41778-3_18}}.

\bibitem{escaleraactivityrecognition}
K.~Nasrollahi, S.~Escalera, P.~Rasti, G.~Anbarjafari, X.~Baro, H.~J. Escalante,
  T.~B. Moeslund, Deep learning based super-resolution for improved action
  recognition, in: 2015 International Conference on Image Processing Theory,
  Tools and Applications (IPTA), 2015, pp. 67--72.
\newblock \href {http://dx.doi.org/10.1109/IPTA.2015.7367098}
  {\path{doi:10.1109/IPTA.2015.7367098}}.

\bibitem{Shotton13}
J.~Shotton, R.~Girshick, A.~Fitzgibbon, T.~Sharp, M.~Cook, M.~Finocchio,
  R.~Moore, P.~Kohli, A.~Criminisi, A.~Kipman, A.~Blake, Efficient human pose
  estimation from single depth images, IEEE Transactions on Pattern Analysis
  and Machine Intelligence 35~(12) (2013) 2821--2840.
\newblock \href
  {http://dx.doi.org/http://doi.ieeecomputersociety.org/10.1109/TPAMI.2012.241}
  {\path{doi:http://doi.ieeecomputersociety.org/10.1109/TPAMI.2012.241}}.

\bibitem{temporalrbm}
I.~Sutskever, G.~E. Hinton, Learning multilevel distributed representations for
  high-dimensional sequences., in: M.~Meila, X.~Shen (Eds.), AISTATS, Vol.~2 of
  JMLR Proceedings, JMLR.org, 2007, pp. 548--555.

\bibitem{originalrbm}
P.~Smolensky, Information processing in dynamical systems: Foundations of
  harmony theory, in: D.~E. Rumelhart, J.~L. McClelland, et~al. (Eds.),
  Parallel Distributed Processing: Volume 1: Foundations, MIT Press, Cambridge,
  1987, pp. 194--281.

\bibitem{taylorcrbmicml}
G.~W. Taylor, G.~E. Hinton, Factored conditional restricted boltzmann machines
  for modeling motion style, in: Proceedings of the 26th Annual International
  Conference on Machine Learning, ICML '09, 2009, pp. 1025--1032.

\bibitem{gwtaylorhdts}
G.~W. Taylor, G.~E. Hinton, S.~T. Roweis, Two distributed-state models for
  generating high-dimensional time series, Journal of Machine Learning Research
  12 (2011) 1025--1068.

\bibitem{ffwcrbmprl}
D.~C. Mocanu, H.~Bou-Ammar, D.~Lowet, K.~Driessens, A.~Liotta, G.~Weiss,
  K.~Tuyls, Factored four way conditional restricted boltzmann machines for
  activity recognition, Pattern Recognition Letters\href
  {http://dx.doi.org/http://dx.doi.org/10.1016/j.patrec.2015.01.013}
  {\path{doi:http://dx.doi.org/10.1016/j.patrec.2015.01.013}}.

\bibitem{mocanugenerativereplay}
D.~C. Mocanu, M.~T. Vega, E.~Eaton, P.~Stone, A.~Liotta, Online contrastive
  divergence with generative replay: Experience replay without storing data,
  CoRR abs/1610.05555.

\bibitem{bengiodl}
Y.~Bengio, Learning deep architectures for ai, Found. Trends Mach. Learn. 2~(1)
  (2009) 1--127.

\bibitem{mocanumljxbm}
D.~C. Mocanu, E.~Mocanu, P.~H. Nguyen, M.~Gibescu, A.~Liotta, A topological
  insight into restricted boltzmann machines, Machine Learning 104~(2) (2016)
  243--270.
\newblock \href {http://dx.doi.org/10.1007/s10994-016-5570-z}
  {\path{doi:10.1007/s10994-016-5570-z}}.

\bibitem{hintoncd}
G.~E. Hinton, {Training Products of Experts by Minimizing Contrastive
  Divergence}, Neural Computation 14~(8) (2002) 1771--1800.

\bibitem{Ponti2017470}
M.~Ponti, J.~Kittler, M.~Riva, T.~de~Campos, C.~Zor, A decision cognizant
  kullback–leibler divergence, Pattern Recognition 61 (2017) 470 -- 478.
\newblock \href
  {http://dx.doi.org/http://dx.doi.org/10.1016/j.patcog.2016.08.018}
  {\path{doi:http://dx.doi.org/10.1016/j.patcog.2016.08.018}}.

\bibitem{Elaiwat2016152}
S.~Elaiwat, M.~Bennamoun, F.~Boussaid, A spatio-temporal rbm-based model for
  facial expression recognition, Pattern Recognition 49 (2016) 152 -- 161.
\newblock \href
  {http://dx.doi.org/http://dx.doi.org/10.1016/j.patcog.2015.07.006}
  {\path{doi:http://dx.doi.org/10.1016/j.patcog.2015.07.006}}.

\bibitem{modellingjointdensities}
G.~Hinton, M.~Pollefeys, J.~Susskind, R.~Memisevic, Modeling the joint density
  of two images under a variety of transformations, 2013 IEEE Conference on
  Computer Vision and Pattern Recognition 00~(undefined) (2011) 2793--2800.
\newblock \href
  {http://dx.doi.org/doi.ieeecomputersociety.org/10.1109/CVPR.2011.5995541}
  {\path{doi:doi.ieeecomputersociety.org/10.1109/CVPR.2011.5995541}}.

\bibitem{hintontrain}
G.~Hinton, A practical guide to training restricted boltzmann machines, in:
  Neural Networks: Tricks of the Trade, Vol. 7700 of Lecture Notes in Computer
  Science, Springer, 2012, pp. 599--619.
\newblock \href {http://dx.doi.org/10.1007/978-3-642-35289-8_32}
  {\path{doi:10.1007/978-3-642-35289-8_32}}.

\bibitem{vapniksvm}
C.~Cortes, V.~Vapnik, {Support-Vector Networks}, Mach. Learn. 20~(3) (1995)
  273--297.

\bibitem{roc}
T.~Fawcett, {An introduction to ROC analysis}, Pattern Recognition Letters
  27~(8) (2006) 861--874.

\bibitem{IonescuSminchisescu11}
C.~S. Catalin~Ionescu, Fuxin~Li, Latent structured models for human pose
  estimation, in: International Conference on Computer Vision, 2011.

\bibitem{h36m_pami}
C.~Ionescu, D.~Papava, V.~Olaru, C.~Sminchisescu, Human3.6m: Large scale
  datasets and predictive methods for 3d human sensing in natural environments,
  IEEE Transactions on Pattern Analysis and Machine Intelligence 36~(7) (2014)
  1325--1339.

\end{thebibliography}

\end{document}